\documentclass[runningheads]{llncs}

\usepackage[T1]{fontenc}
\usepackage{graphicx}
\usepackage[labelformat=simple]{subcaption}
	
\usepackage{amsmath,amssymb}
\usepackage{algorithm}
\usepackage[dvipsnames]{xcolor}
	\definecolor{kwgreen}{rgb}{0.00,0.50,0.00}
\usepackage{listings}
\lstdefinelanguage{Julia}%
  {morekeywords={abstract,break,case,catch,const,continue,do,else,elseif,%
      end,export,false,for,function,immutable,import,importall,if,in,%
      macro,module,otherwise,quote,return,switch,true,try,type,typealias,%
      using,while,let},%
   sensitive=true,%
   alsoother={$},%
   morecomment=[l]\#,%
   morecomment=[n]{\#=}{=\#},%
   morestring=[s]{"}{"},%
   morestring=[m]{'}{'},%
}[keywords,comments,strings]%
\usepackage{xspace}
\usepackage{hyperref}

\newcommand{\figref}[1]{Fig.~\ref{fig:#1}\xspace}
\newcommand{\exampleend}{\unskip\nobreak\hfill\ensuremath{\lhd}}
\newcommand{\R}{\ensuremath{\mathbb{R}}\xspace}
\newcommand{\NN}{\ensuremath{N}\xspace}
\newcommand{\layer}{\ensuremath{\ell}\xspace}
\newcommand{\act}{\ensuremath{\alpha}\xspace}
\newcommand{\relu}{\ensuremath{\rho}\xspace}
\newcommand{\lrelu}[1]{\ensuremath{\rho_{#1}^\ell}\xspace}
\newcommand{\sigmoid}{\ensuremath{\sigma}\xspace}
\newcommand{\set}[1]{\ensuremath{\mathcal{#1}}\xspace}
\newcommand{\X}{\set{X}\xspace}
\newcommand{\Y}{\set{Y}\xspace}
\newcommand{\C}{\set{C}\xspace}
\renewcommand{\P}{\set{P}\xspace}
\newcommand{\Part}{\ensuremath{\Pi}\xspace}
\newcommand{\Oof}{\ensuremath{\mathcal{O}}\xspace}

\title{The inverse problem for neural networks}
\author{
Marcelo Forets\inst{1}\orcidID{0000-0002-9831-7801}
\and \\
Christian Schilling\inst{2}\orcidID{0000-0003-3658-1065}
}
\institute{
Universidad de la Rep\'{u}blica, Uruguay
\and
Aalborg University, Denmark
}

\begin{document}

\maketitle

\begin{abstract}
We study the problem of computing the preimage of a set under a neural network with piecewise-affine activation functions.
We recall an old result that the preimage of a polyhedral set is again a union of polyhedral sets and can be effectively computed.
We show several applications of computing the preimage for analysis and interpretability of neural networks.

\end{abstract}

\section{Introduction}

We study the inverse problem for neural networks.
That is, given a neural network $\NN$ and a set $Y$ of outputs, we want to compute the preimage, i.e., we want to find all inputs $x$ such that their image under the network is in $Y$ ($\NN(x) \in Y$).

Computing the preimage has at least two motivations.
First, it can be used for specification mining.
Besides the obvious benefit of actually obtaining a potential specification, the main application is in interpretability, explaining the function the neural network encodes.
This has indeed been investigated before (under the name \emph{rule extraction})~\cite{Thrun94,Maire99,BreutelMH03}.
As a second motivation, the preimage is also useful if a specification is known.
In that case we can analyze whether the specification holds, e.g., whether there exist inputs leading to a set of outputs.

In this paper, we give a complete picture of the preimage computation for piecewise-affine neural networks.
For this class, the preimage of a polyhedral set is again a union of (potentially exponentially many) polyhedral sets and can be effectively computed using linear programming.
This result has already been obtained in the past~\cite{Maire99}.
However, that line of work seems to not be well known, a potential reason being that, at the time, piecewise-affine activation functions were not used.
That work is indeed only concerned with a piecewise-affine approximation of the then-common sigmoid activation functions.
Nowadays, piecewise-affine activation functions are most widely used in practice, and have also been extensively investigated in the formal-methods community.
We believe that the community has fruitful applications for the preimage, and the purpose of this paper is to bring these results to the community's attention.
We also show a potential use case in interpretability and two extensions by approximation and combination with forward-image computation.

\subsection{Related work}

The traditional take on the inverse problem for neural networks in the machine-learning community has particularly been studied for image classifiers, where the task is to either highlight which neurons~\cite{MahendranV15} or input pixels~\cite{SimonyanVZ13,ZeilerF14} were most influential in the decision.
However, these approaches do not compute the actual inverse but rather implement heuristics to map to one particular input.
Many such techniques have been demonstrated to be misleading~\cite{SixtGL20}.
For instance, the technique in~\cite{ZeilerF14} remembers the inputs to a max-pooling layer, which otherwise cannot be inverted.
A related concept is the autoencoder~\cite{KingmaW13}, where the decoder maps the output of the encoder back to the input; however, both maps are learned and thus there is no guarantee that the decoder inverts the encoder.

Another direction of inverse computation in neural networks is known as adversarial attacks~\cite{GoodfellowSS14}.
The motivation is to find inputs that drive a classifier to a misclassification.
For that, an input is mildly perturbed such that it crosses the network's decision boundary.

In abstract interpretation~\cite{CousotC77}, the abstraction function maps an input to an abstract domain, and the concretization function maps back to the (set of) concrete values.
Abstract interpretation has also been used for (forward) set propagation in neural networks with abstract domains such as intervals~\cite{Thrun94,WangPWYJ18}, polytopes~\cite{YangJT0HP21}, zonotopes~\cite{SinghGMPV18,SchillingFG22}, and polynomial zonotopes~\cite{KochdumperSAB23}.
A related approach uses polygons: given a classifier with two-dimensional input domain, the approach partitions the input domain into the preimages of each class~\cite{SotoudehTT23}.

Maire~\cite{Maire99} proposed an algorithm to compute the preimage of a union of polyhedra.
The algorithm applies to arbitrary activation functions, but only computes an approximate solution in general.
To improve scalability, Breutel and Maire use an additional approximation step based on nonlinear optimization~\cite{BreutelMH03}.
Our goal is the exact computation of the preimage for networks with piecewise-affine activations.
Although very close to this work, the above approaches assume a bounded input domain, bounded activation space, or surjective activations, which excludes the (nowadays) widely used rectified linear unit.

Computing the inverse of a function under a set is also known as set inversion, which has been mainly studied in the context of interval constraint propagation, with applications in robotics, control, or parameter estimation~\cite{Jaulin21}.
To reduce the approximation error, one can apply iterative forward-backward contractors~\cite{ChabertJ09}.
We show such an experiment later.
Thrun applied this idea to neural networks with bounded input and output domains as well as activation spaces~\cite{Thrun94}.

Recently, a number of works study backward reachability in discrete-time neural-network control systems~\cite{RoberEH22,BakT22,EverettBO23,RoberEZH23,KothaBKDZ23}.
These approaches do not seem to be aware of the above-mentioned old works, and discuss strategies to compute an approximation of the preimage.
This is due to the intractability of computing the exact preimage.

\section{The preimage of a piecewise-affine neural network}

In this section, we describe an algorithm to compute the preimage of a union of polyhedra under a neural network with piecewise-affine activations.

\subsection{Preliminaries}

A function $f : \R^m \to \R^n$ is affine if $f(x) = Cx + d$ for some matrix $C \in \R^{n \times m}$ and vector $d \in \R^n$.
The function $f$ is piecewise affine if there exists a partitioning of $\R^m$ such that $f$ is affine in each partition.

A (deep) neural network (DNN)\footnote{Typically, \emph{deep} neural networks are neural networks with multiple hidden layers. Here we do not make this distinction and always use the term \emph{DNN}.} $\NN$ comprises $k$ layers $\layer_i$ that are sequentially composed such that $\NN = \layer_k \circ \dots \circ \layer_1$.
Each layer $\layer_i$ applies an affine map of appropriate dimensions, followed by an activation function, written $\layer_i(x) = \act_i(W_i x + b_i)$.
Activation functions are one-dimensional maps $\act_i : \R \to \R$, and are extended componentwise to vectors.
Some common piecewise-affine activation functions are the identity, the rectified linear unit (ReLU), $\relu(x) = \max(x, 0)$, and the leaky ReLU, which is a parametric generalization with $a \geq 0$ defined as
\[
	\lrelu{a}(x) = \begin{cases} x & x > 0 \\ ax & x \leq 0. \end{cases}
\]

An $n$-dimensional half-space (or linear constraint) is characterized by a vector $c \in \R^n$ and a scalar $d \in \R$ and represents the set $\{x : c^T x \leq d\}$.
A polyhedron is an intersection of half-spaces.
A convenient way to write a polyhedron is in the matrix-vector form $Cx \leq d$, where the $i$-th row of $C$ and $d$ correspond to the $i$-th half-space.
A polytope is a bounded polyhedron.

Given a function (e.g., a DNN) $f : \R^m \to \R^n$, the \emph{image} of a set $\X \subseteq \R^m$ is $f(\X) = \{f(x) : x \in \X\} \subseteq \R^n$.
Analogously, the \emph{preimage} of a set $\Y \subseteq \R^n$ is $f^{-1}(\Y) = \{x : f(x) \in \Y\} \subseteq \R^m$.
For nonempty sets and injective $g : \R^m \to \R^n$ we note the following facts~\cite{Halmos60,Lee10}:
\begin{align}
	\X &\subseteq f^{-1}(f(\X)) \label{eq:preimage_image_general} \\
	\X &= g^{-1}(g(\X)) \label{eq:preimage_image} \\
	f(\X_1 \cup \X_2) &= f(\X_1) \cup f(\X_2) \label{eq:image_cup} \\
	f^{-1}(\Y_1 \cup \Y_2) &= f^{-1}(\Y_1) \cup f^{-1}(\Y_2) \label{eq:preimage_cup} \\
	f(\X) \cap \Y &= f(\X \cap f^{-1}(\Y)) \label{eq:mixed_cap}
\end{align}

The inclusion in Eq.~\eqref{eq:preimage_image_general} is generally strict.
In particular, for the class of DNNs considered here, the image of a (bounded) polytope $\X$ is a union of polytopes, but the preimage of a nonempty polytope (even if consisting of a single element) is a union of (unbounded) polyhedra.
Because of the layer-wise architecture of DNNs, the image can be computed by iteratively applying each affine map and activation function.
Below we describe how to compute the preimage in the same fashion, most of which has been described in prior work~\cite{Maire99}.

\subsection{Inverse affine map}

Given sets $\X \subseteq \R^m$ and $\Y \subseteq \R^n$, a matrix $W \in \R^{n \times m}$, and a vector $b \in \R^n$, for the affine map $f(x) = W x + b$, the image of $\X$ is $f(\X) = \{f(x) : x \in \X\} \subseteq \R^n$ and the preimage of $\Y$ is $f^{-1}(\Y) = \{x : W x + b \in \Y\} \subseteq \R^n$.
For a polyhedron $\Y$ written as $Cy \leq d$, we can write the preimage as follows:
\begin{equation}\label{eq:inv_aff}
	f^{-1}(\Y) = \{x : C(Wx + b) \leq d\} = \{x : CWx \leq d - Cb\}
\end{equation}

Note that some of the resulting linear constraints may be redundant or contradictory.
In the latter case, the preimage is empty.

\begin{example}\label{ex:inv_aff}
	For the affine map $f(x) = \begin{pmatrix} -0.46 & 0.32 \end{pmatrix} x + 2$ and the interval $\Y = [2, 3]$, we get the infinite band $f^{-1}(\Y) = \{x \in \R^2 : 0 \leq \begin{pmatrix} -0.46 & 0.32 \end{pmatrix} x \leq 1\}$.
	\exampleend
\end{example}

\subsection{Inverse piecewise-affine activation function}

Given sets $\X, \Y \subseteq \R^n$ and a piecewise-affine activation function $\act$, the image of $\X$ can be computed as follows.
Consider the partitioning $\Part = \bigcup_j \P_j$ of $\R^n$ into the different domains $\P_j$ in which $\act$ is an affine function $\act_j$.
(Observe that $\act_j$ has a diagonal matrix due to componentwise application.)
Then we can apply the corresponding affine map to the intersection of $\X$ with each $\P_j$:
\[
	\act(\X)
	= \act\big(\bigcup_j \P_j \cap \X\big)
	\stackrel{\eqref{eq:image_cup}}{=} \bigcup_j \act(\P_j \cap \X)
	= \bigcup_j \act_j(\P_j \cap \X)
\]

\begin{example}\label{ex:forward_relu}
	Consider the ReLU function $\rho$ for $n = 2$.
	The partitioning \Part consists of four regions with the corresponding affine maps being modified identity matrices, where diagonal entry $i$ is zero if the $i$-th dimension is non-positive:
	\begin{align*}
		\P_1 = x_1 > 0 \land x_2 > 0 &\qquad \rho_1(x) = \begin{pmatrix} 1 & 0 \\ 0 & 1 \end{pmatrix} x \\
		\P_2 = x_1 \leq 0 \land x_2 > 0 &\qquad \rho_2(x) = \begin{pmatrix} 0 & 0 \\ 0 & 1 \end{pmatrix} x \\
		\P_3 = x_1 > 0 \land x_2 \leq 0 &\qquad \rho_3(x) = \begin{pmatrix} 1 & 0 \\ 0 & 0 \end{pmatrix} x \\
		\P_4 = x_1 \leq 0 \land x_2 \leq 0 &\qquad \rho_4(x) = \begin{pmatrix} 0 & 0 \\ 0 & 0 \end{pmatrix} x
	\end{align*}

	Thus we get:
	\begin{align*}
		\rho(\X) = \rho_1(\P_1 \cap \X) \cup \rho_2(\P_2 \cap \X) \cup \rho_3(\P_3 \cap \X) \cup \rho_4(\P_4 \cap \X) \tag*{\exampleend}
	\end{align*}
\end{example}

The preimage can be computed in an analogous way, using that $\bigcup_j \alpha_j(\P_j) = \alpha(\R^n)$.
Note that, while the sets $\P_j$ partition $\R^n$, the sets $\alpha_j(\P_j)$ do not necessarily partition $\R^n$ (only if $\alpha$ is injective).
\begin{align}
	\act^{-1}(\Y)
	&= \act^{-1}\big(\bigcup_j \act_j(\P_j) \cap \Y\big) \notag \\
	&\stackrel{\eqref{eq:preimage_cup}}{=} \bigcup_j \act_j^{-1}(\act_j(\P_j) \cap \Y) \label{eq:preimage_inv_simple} \\
	&\stackrel{\eqref{eq:mixed_cap}}{=} \bigcup_j \act_j^{-1} (\act_j(\P_j \cap \act_j^{-1}(\Y))) \label{eq:preimage_inv}
\end{align}

One can implement a check for Eq.~\eqref{eq:preimage_inv_simple} directly, which is done in~\cite{Maire99}.
However, there are two issues with Eq.~\eqref{eq:preimage_inv_simple}.

(1)~Some of the resulting linear constraints may be redundant.
This is also noted in~\cite{Maire99}.
For example, in the case of ReLU and assuming that $\Y$ contains the origin, the negative orthant is part of the preimage and can be represented by $n$ linear constraints; instead, the equation suggests to compute the image of the negative orthant (which is the set just containing the origin), which already requires at least $n+1$ linear constraints, then computes the intersection with $\Y$, and finally computes the preimage.

(2)~Instead of computing the preimage of the intersection of $\act_j(\P_j)$ with $\Y$, it may be more efficient to compute the preimage of $\Y$ first.
The reason is that the preimage computation is more efficient for certain set representations, but the intersection may remove the structure from the set.

Eq.~\eqref{eq:preimage_inv} is an attempt to mitigate these issues.
While this expression looks more complicated, we can simplify it further by considering two cases.
Recall that $\act_j$ is an affine activation function, i.e., it is applied componentwise, and observe that affine functions $\act_j: \R \to \R$ are either injective or constant.
In the further case we can directly apply Eq.~\eqref{eq:preimage_image}.
\begin{equation}\label{eq:preimage_inv_injective}
	\act_j^{-1} (\act_j(\P_j \cap \act_j^{-1}(\Y))) = \P_j \cap \act_j^{-1}(\Y)
\end{equation}

If $\act_j$ is constant such that $\act_j(x) = c$ for all $x \in \P_j$, then for any $\X$ we have
\[
	\act_j(\P_j \cap \X) = \begin{cases}
		\{c\} & \P_j \cap \X \neq \emptyset \\
		\emptyset & \P_j \cap \X = \emptyset.
	\end{cases}
\]
Since $\act_j^{-1}(\Y) = \P_j$ if $c \in \Y$ and $\act_j^{-1}(\Y) = \emptyset$ if $c \notin \Y$, we get:
\begin{equation}\label{eq:preimage_inv_constant}
	\act_j^{-1} (\act_j(\P_j \cap \act_j^{-1}(\Y))) = \begin{cases}
		\P_j & c \in \Y \\
		\emptyset & c \notin \Y
	\end{cases}
\end{equation}

\begin{example}
	We continue with the infinite band from Example~\ref{ex:inv_aff}, which we rename to $\Y$, and compute the inverse under ReLU, i.e., $\relu^{-1}(\Y)$.
	Since we have two dimensions, we have the four partitions from Example~\ref{ex:forward_relu}.
	There are two cases.
	Case~1 implements Eq.~\eqref{eq:preimage_inv_injective}, which applies to $\P_1$, $\P_2$, and $\P_3$.
	Case~2 implements Eq.~\eqref{eq:preimage_inv_constant}, which applies to $\P_4$.
	Since the ReLU constant is $c = (0, 0)^T$, we get $\X_4 := \P_4$ because $c \in \Y$.
	Overall the result is $\X_1 \cup \X_2 \cup \X_3 \cup \X_4$, where
	\begin{align*}
		\X_1 &:= \P_1 \cap \rho_1^{-1}(\Y) = \{x \in \R^2 : 0 \leq \begin{pmatrix} -0.46 & 0.32 \end{pmatrix} x \leq 1 \land x_1 > 0\} \\
		\X_2 &:= \P_2 \cap \rho_2^{-1}(\Y) = \{x \in \R^2 : x_1 \leq 0 \land 0 \leq x_2 \leq 3.125\} \\
		\X_3 &:= \P_3 \cap \rho_3^{-1}(\Y) = \emptyset \\
		\X_4 &:= \P_4 = \{x \in \R^2 : x_1 \leq 0 \land x_2 \leq 0\}. \tag*{\exampleend}
	\end{align*}
\end{example}

\subsection{Inverse piecewise-affine DNN}

\begin{algorithm}[t!]
\caption{Preimage computation for piecewise-affine neural networks}
\label{algo:preimage}
\begin{lstlisting}
# inverse neural network
function preimage(Z, |$\NN$|)
    let |$\layer$|_1, ..., |$\layer$|_k be the layers of |$\NN$|
    for |$\layer$| in |$\layer$|_k, ..., |$\layer$|_1
        let W, b, |$\alpha$| be the components of |$\layer$|
        Y = preimage(Z, |$\alpha$|)
        X = preimage(Y, W, b)
        Z = X
    end
    return X
end

# inverse piecewise-affine activation function
function preimage(Z, |$\alpha$|)
    Y = |$\emptyset$|
    let n be the dimension of Z
    # loop over partitions and corresponding affine functions
    for (Pj, |$\alpha$|j) in PWA_partitioning(|$\alpha$|, n)
        let A_|$\alpha$|j * x = b_|$\alpha$|j be the affine representation of |$\alpha$|j
        if iszero(A_|$\alpha$|j) # constant case, Eq. (10)
            if b_|$\alpha$|j |$\in$| Z
                Y = Y |$\cup$| Pj
            end
        else  # injective case, Eq. (9)
            Q = preimage(Z, A_|$\alpha$|j, b_|$\alpha$|j)
            Y = Y |$\cup$| (Pj |$\cap$| Q)
        end
    end
    return Y
end

# inverse affine map, Eq. (6)
function preimage(Y, W, b)
    let C |$\leq$| d be the constraints of Y
    M = C * W
    v = d - C * b
    return Polyhedron(M |$\leq$| v)
end
\end{lstlisting}
\end{algorithm}

The procedure for a whole DNN is the straightforward alternation of Eq.~\eqref{eq:preimage_inv_simple} and Eq.~\eqref{eq:inv_aff} for each layer.
In our implementation, instead of Eq.~\eqref{eq:preimage_inv_simple} we use Eq.~\eqref{eq:preimage_inv} with a case distinction for whether to simplify to Eq.~\eqref{eq:preimage_inv_injective} or Eq.~\eqref{eq:preimage_inv_constant}.
The procedure is summarized in Algorithm~\ref{algo:preimage}.

\smallskip

Regarding complexity, given a DNN with $k$ layers each of dimension $n$, it is well-known that the network can map a polyhedron to $\Oof(b^{kn})$ polyhedra, where $b$ is the number of affine pieces in the activation function (e.g., $b = 2$ for ReLU)~\cite{MontufarPCB14}.
The same holds for the preimage.

\begin{example}
	\begin{figure}[t]
		\centering
		\begin{align*}
			\layer_1(x) &= \relu\left(\begin{pmatrix} 0.30 & 0.53 \\ 0.77 & 0.42\end{pmatrix} x + \begin{pmatrix} \phantom{-}0.43 \\ -0.42 \end{pmatrix}\right) &
			\layer_3(x) &= \begin{pmatrix} \phantom{-}0.35 & 0.17 \\ -0.04 & 0.08\end{pmatrix} x + \begin{pmatrix} 0.03 \\ 0.17 \end{pmatrix} \\
			\layer_2(x) &= \relu\left(\begin{pmatrix} 0.17 & -0.07 \\ 0.71 & -0.06\end{pmatrix} x + \begin{pmatrix} -0.01 \\ \phantom{-}0.49 \end{pmatrix}\right) &
			\NN &= \layer_3 \circ \layer_2 \circ \layer_1
		\end{align*}
		\caption{A simple DNN $\NN: \R^2 \to \R^2$ for classification.}
		\label{fig:nn_example}
	\end{figure}
	
	\begin{figure}[t]
		\begin{subfigure}[t]{0.48\textwidth}
			\centering
			\includegraphics[width=\textwidth,keepaspectratio]{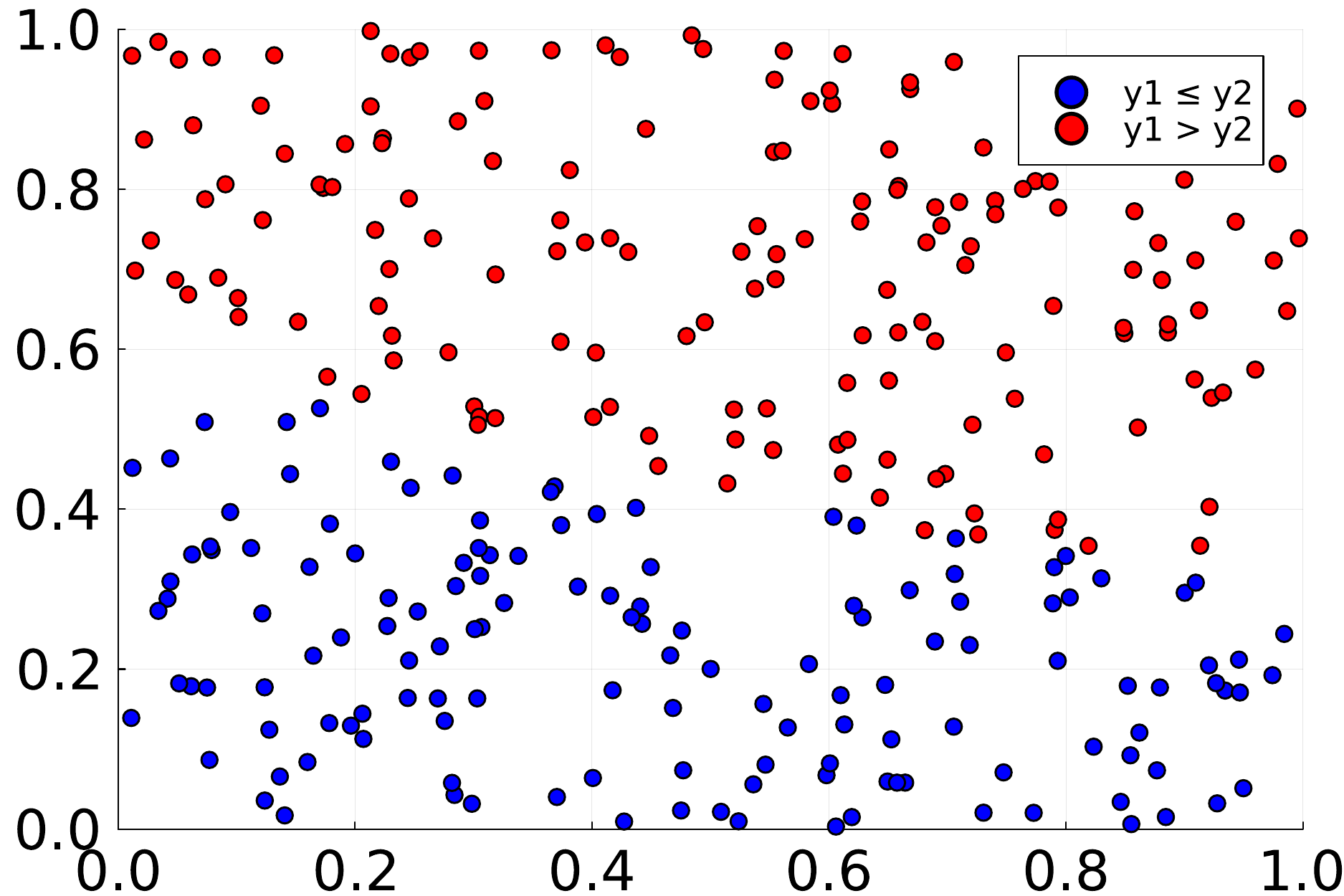}
			\caption{$300$ uniform samples $x \in [0, 1]^2$. Colors show the classification of $y = \NN(x)$.}
			\label{fig:ex_original}
		\end{subfigure}
		\hfill
		\begin{subfigure}[t]{0.48\textwidth}
			\centering
			\includegraphics[width=\textwidth,keepaspectratio]{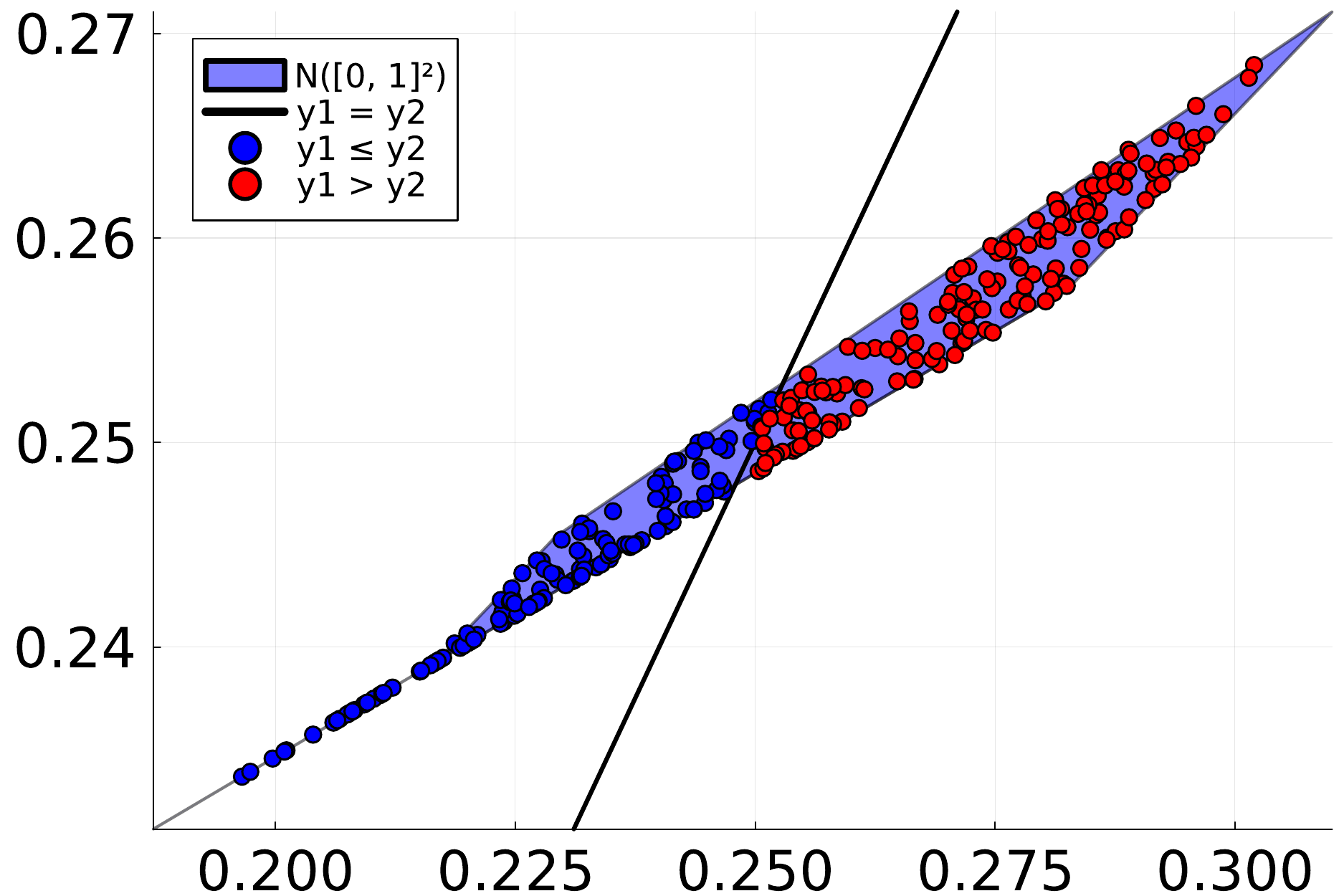}
			\caption{Forward image $\NN([0, 1]^2)$ with samples from \figref{ex_original} under \NN.}
			\label{fig:ex_forward}
		\end{subfigure}
		\caption{Sampled inputs and outputs for the DNN $\NN$ from \figref{nn_example}.}
		\label{fig:example_sampling}
	\end{figure}

	\begin{figure}[t!]
		\begin{subfigure}[t]{0.48\textwidth}
			\centering
			\includegraphics[width=\textwidth,keepaspectratio]{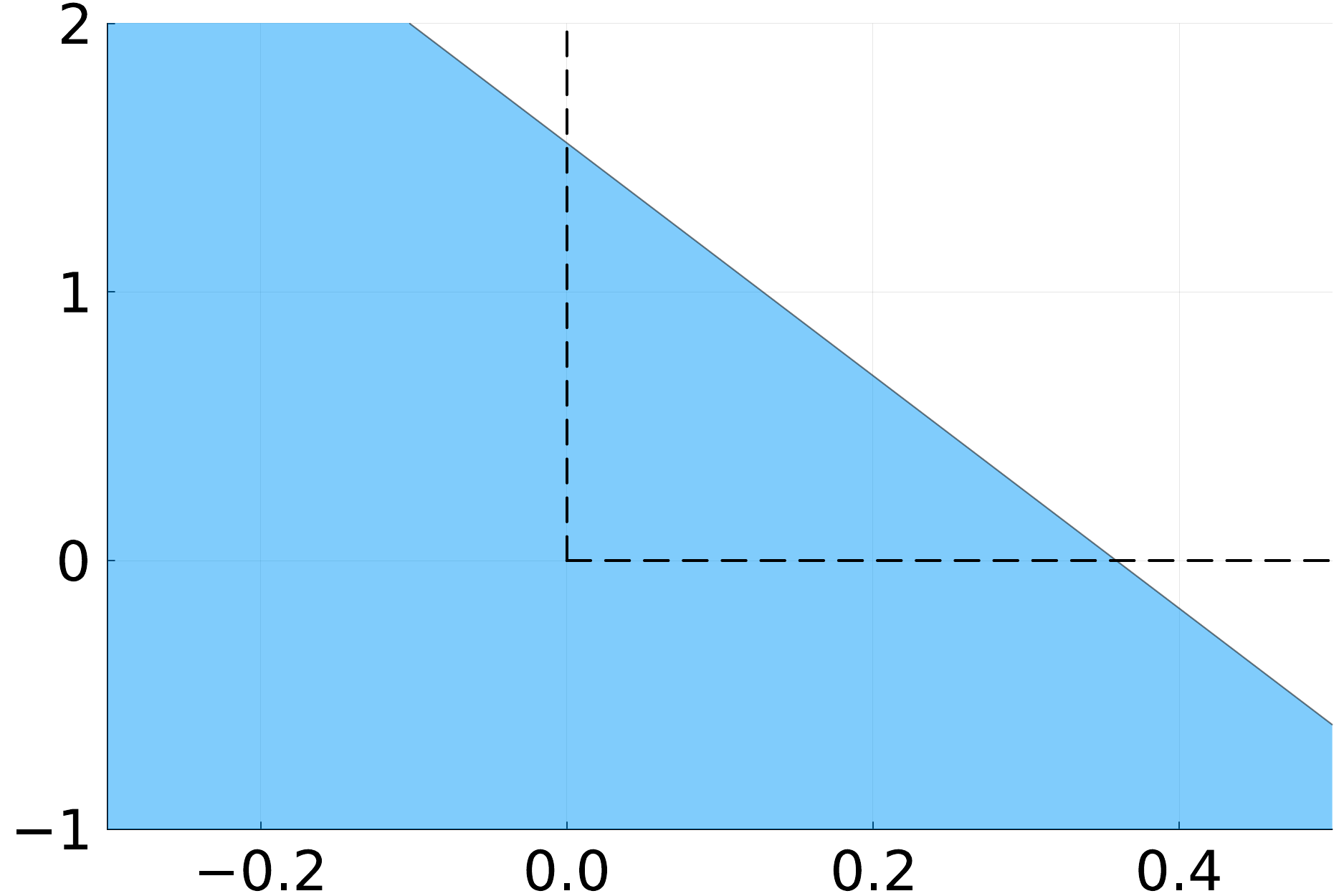}
			\caption{Preimage $\Y_1 = \layer_3^{-1}(\{y : y_1 \leq y_2\})$.}
			\label{fig:ex_backward_l3}
		\end{subfigure}
		\hfill
		\begin{subfigure}[t]{0.48\textwidth}
			\centering
			\includegraphics[width=\textwidth,keepaspectratio]{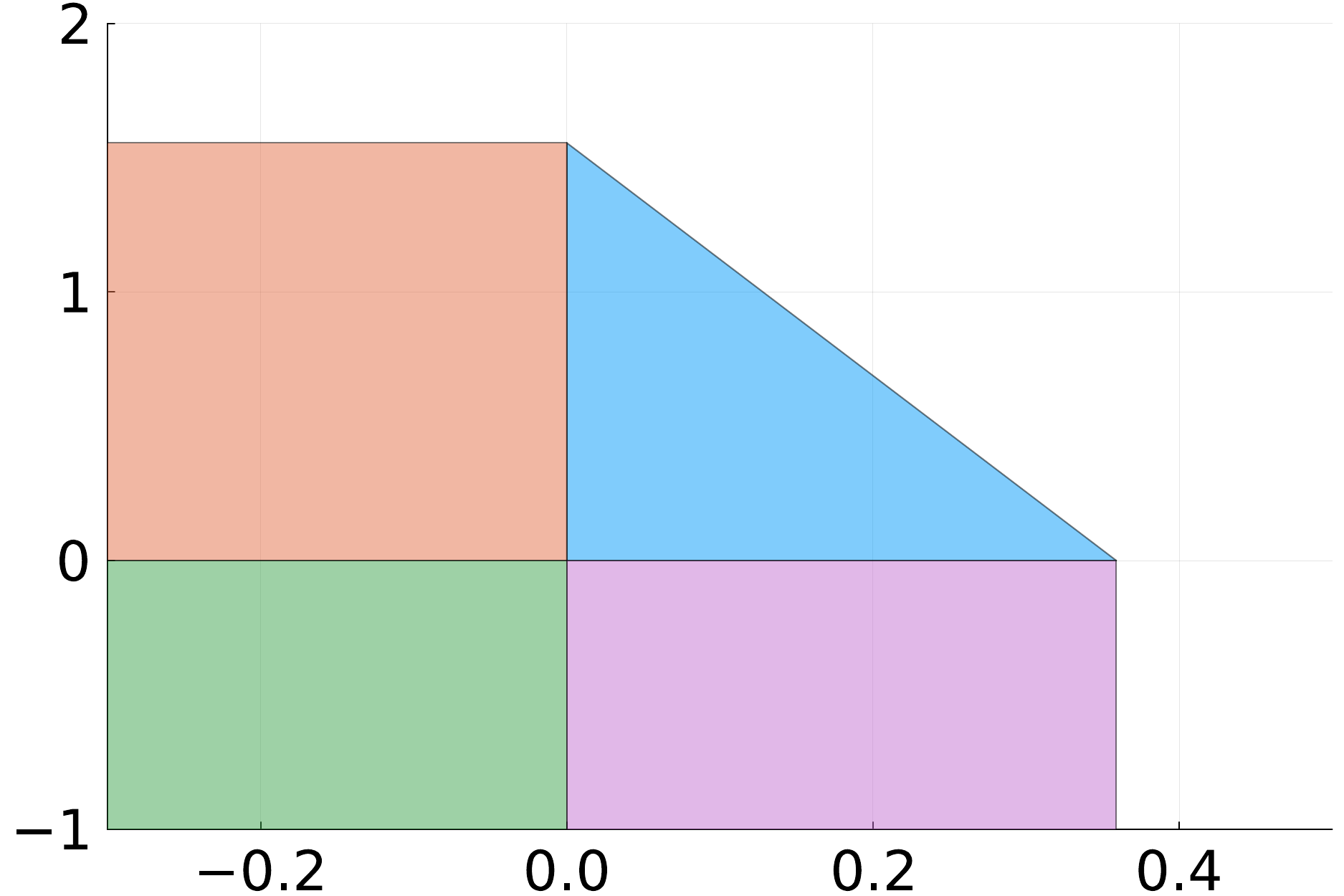}
			\caption{Preimage $\Y_2 = \relu^{-1}(\Y_1)$.}
			\label{fig:ex_backward_l3relu}
		\end{subfigure}
		
		\begin{subfigure}[t]{0.48\textwidth}
			\centering
			\includegraphics[width=\textwidth,keepaspectratio]{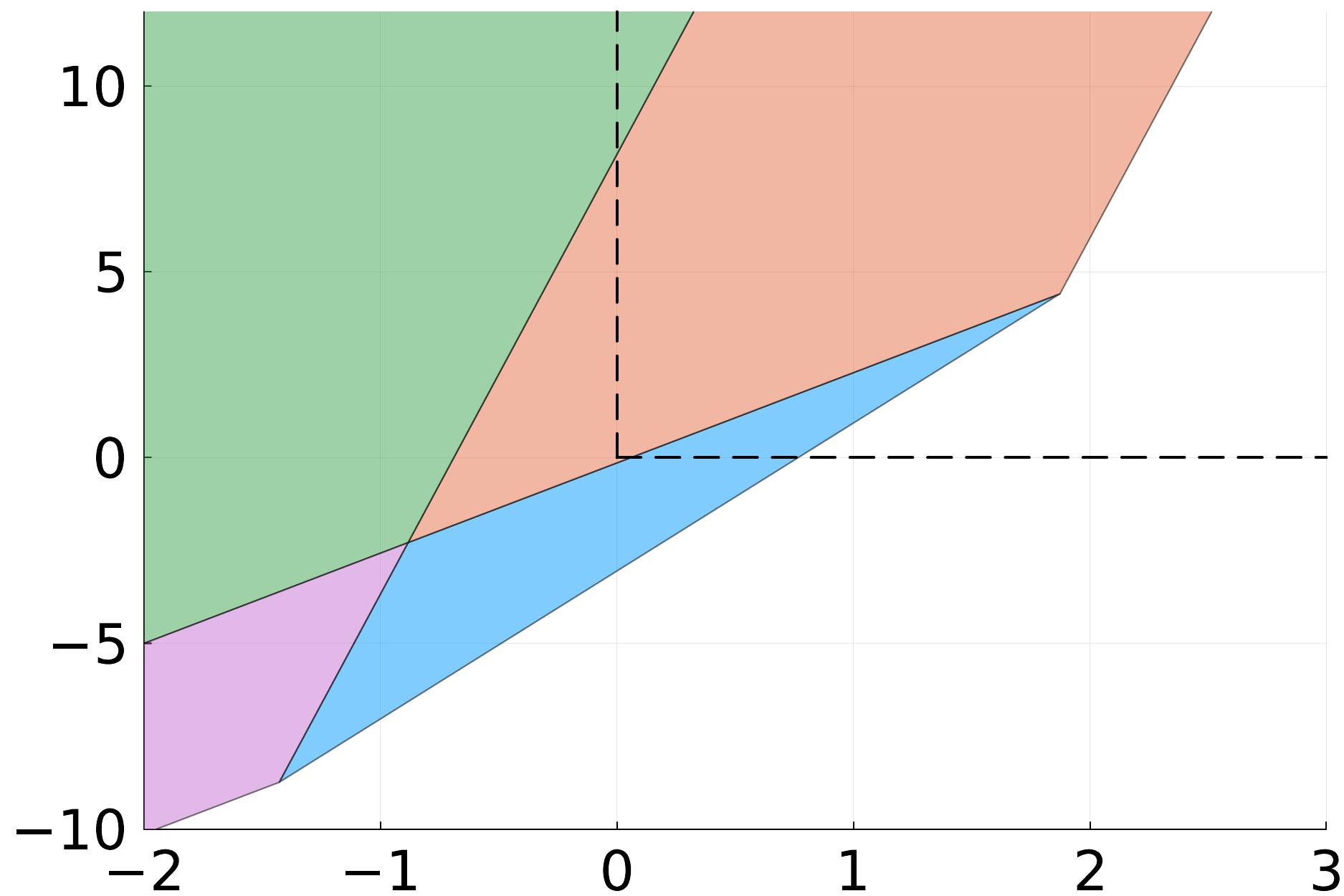}
			\caption{Preimage $\Y_3 = \layer_2^{-1}(\Y_1)$.}
			\label{fig:ex_backward_l2}
		\end{subfigure}
		\hfill
		\begin{subfigure}[t]{0.48\textwidth}
			\centering
			\includegraphics[width=\textwidth,keepaspectratio]{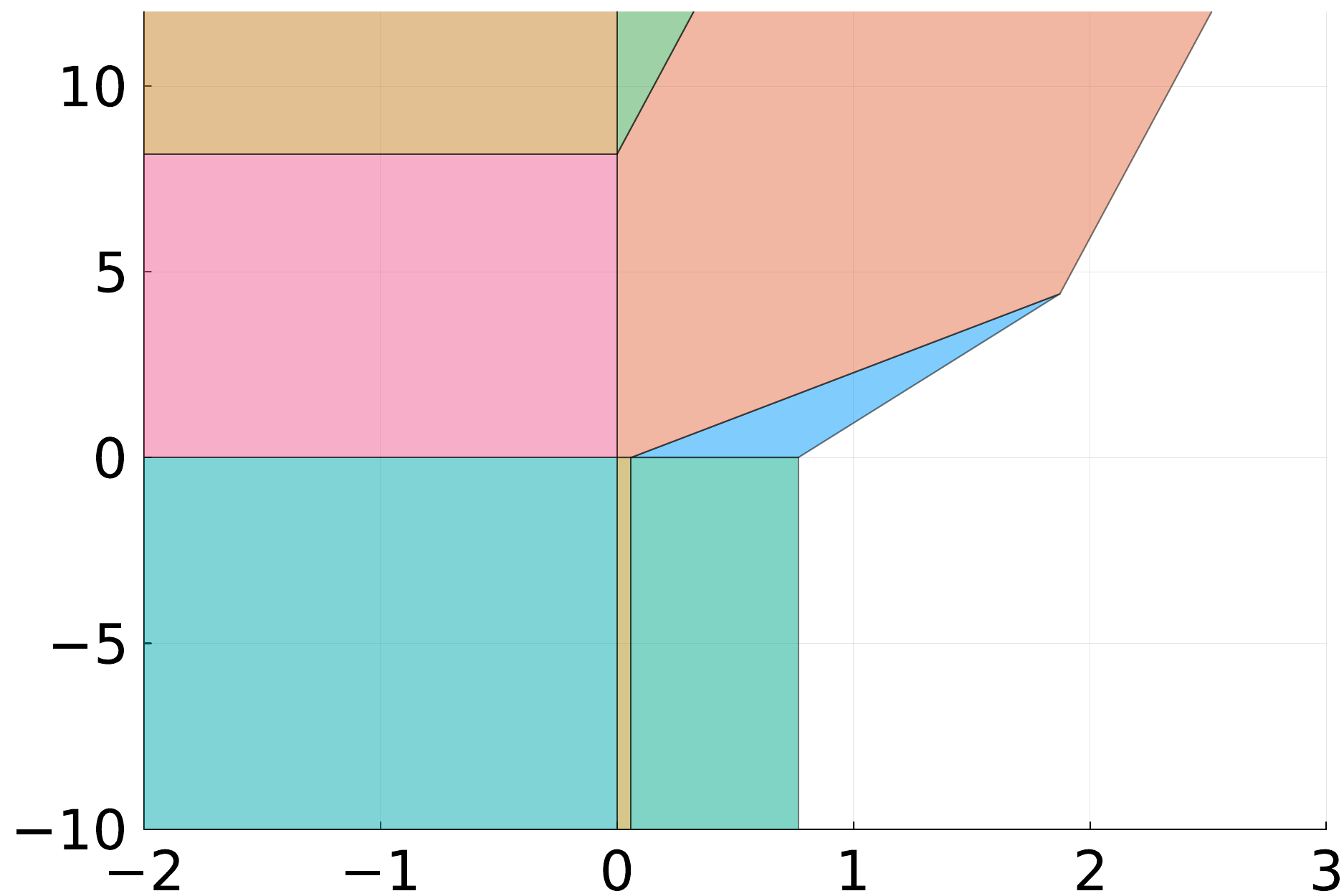}
			\caption{Preimage $\Y_4 = \relu^{-1}(\Y_3)$.}
			\label{fig:ex_backward_l2relu}
		\end{subfigure}
		
		\begin{subfigure}[t]{0.48\textwidth}
			\centering
			\includegraphics[width=\textwidth,keepaspectratio]{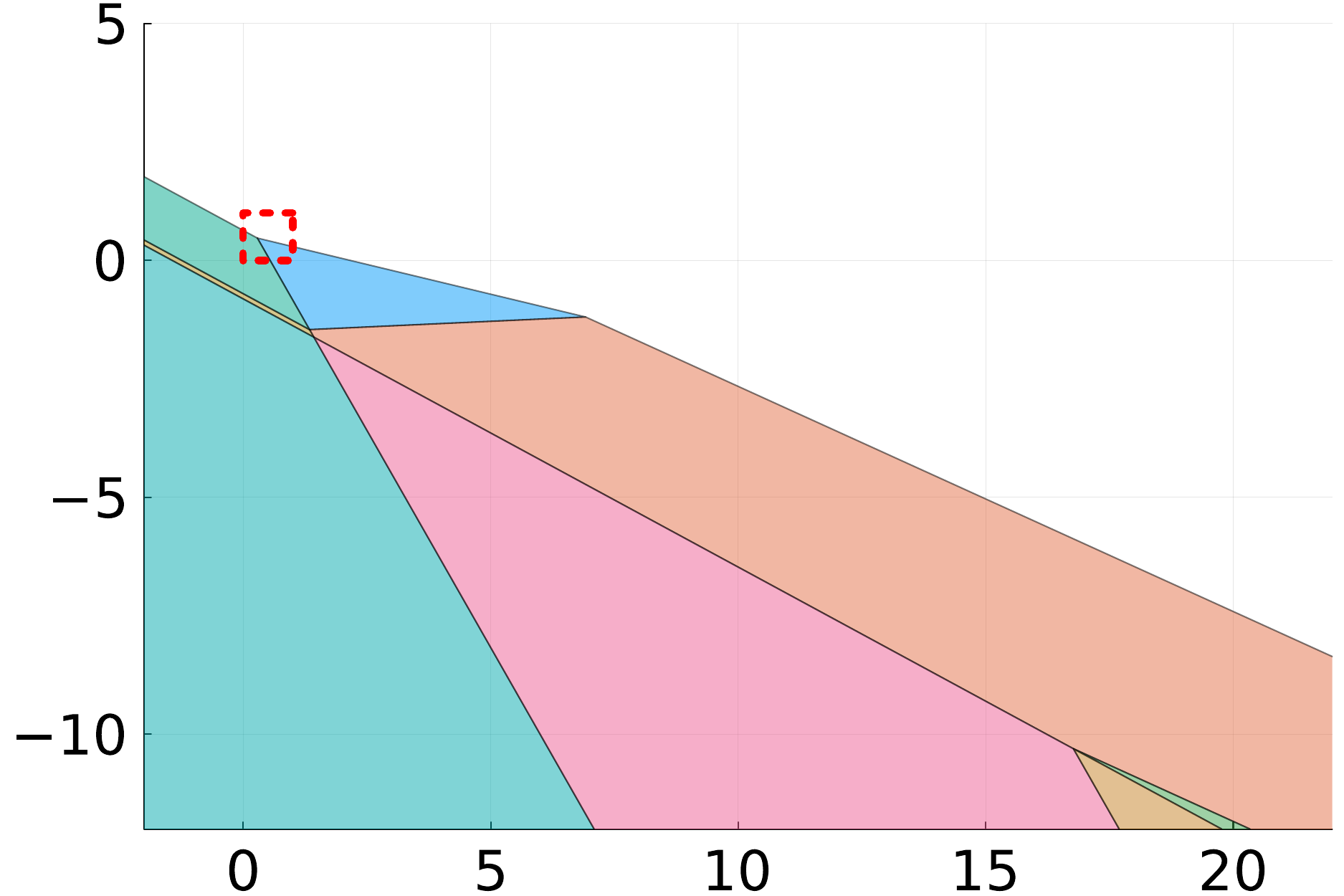}
			\caption{Preimage $\Y_5 = \layer_1^{-1}(\Y_3) = \NN^{-1}(\Y_0)$ with the original domain $[0, 1]^2$ in red.}
			\label{fig:ex_backward_l1}
		\end{subfigure}
		\hfill
		\begin{subfigure}[t]{0.48\textwidth}
			\centering
			\includegraphics[width=\textwidth,keepaspectratio]{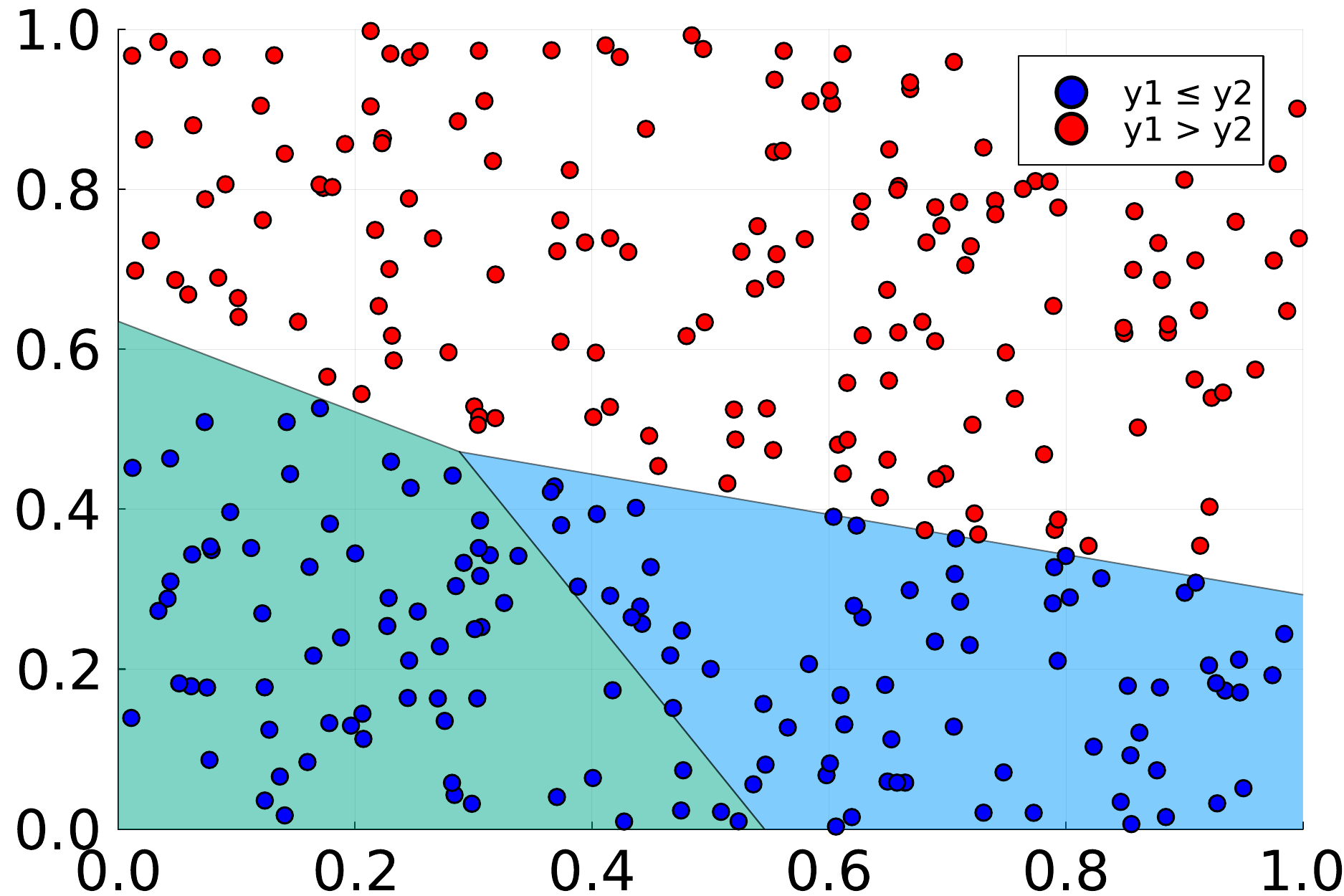}
			\caption{Preimage $\NN^{-1}(\Y_0)$ added to \figref{ex_original}.}
			\label{fig:ex_backward}
		\end{subfigure}
		\caption{Complete example for the DNN $\NN$ from \figref{nn_example}.
		The black dashed lines indicate the codomain of the ReLU activation function.}
		\label{fig:example}
	\end{figure}

	We consider the DNN $\NN$ from \figref{nn_example}.
	Each layer is two-dimensional for convenience of plotting.
	We view the network as a classifier based on which of the two output values is larger.
	To provide some intuition, \figref{ex_original} shows samples in the input domain $[0, 1]^2$ with their associated class.
	\figref{ex_forward} shows the (forward) image of the domain, $\NN([0, 1]^2)$, which is a union of polytopes and can be computed by standard (forward) set propagation.
	The figure also shows the classification boundary (black diagonal).
	
	Now we apply the algorithm to compute the preimage.
	\figref{example} shows different snapshots of the algorithm.
	We compute the preimage of the set $\C_2 = \{y : y_1 \leq y_2\}$ (a half-space) of all inputs that classify as class 2 (blue dots).
	Since the last layer ($\layer_3$) has an identity activation, the preimage $\Y_1 = \layer_3^{-1}(\C_2)$ is just the inverse affine map.
	Since this map is invertible here, the preimage is just another half-space (\figref{ex_backward_l3}).
	Next we compute the preimage under the ReLU activation (\figref{ex_backward_l3relu}).
	The nonnegative part of $\Y_1$ remains, together with the negative extensions.
	After computing the preimage under the next affine map (\figref{ex_backward_l2}) we see that the purple set will get removed with the next ReLU operation.
	While the previous sets could have been simplified to one polyhedron, the next ReLU preimage (\figref{ex_backward_l2relu}) is non-convex.
	Finally, we compute the preimage under the last affine map (\figref{ex_backward_l1}).
	The box in red marks the original domain $[0, 1]^2$.
	\figref{ex_backward} shows the preimage clipped to this domain.
	\exampleend
\end{example}

\section{Applications and extensions}

We implemented the above-described algorithm in the reachability framework JuliaReach~\cite{BogomolovFFPS19}, particularly in the set library LazySets~\cite{ForetsS21b}.
In this section, we report on several experiments and discuss potential extensions.
The code to repeat these experiments is available online~\cite{RE}.

\subsection{Interpretability}

\begin{figure}[t]
	\centering
	\begin{align*}
		\layer_1(x) &= \relu\left(\begin{pmatrix} \phantom{-}0.63 \\ \phantom{-}0.17 \\ -0.79 \end{pmatrix} x + \begin{pmatrix} \phantom{-}0.06 \\ -2.25 \\ -2.27 \end{pmatrix}\right) &
		\layer_3(x) &= \begin{pmatrix} 1.92 & 1.01 & 0.83 \end{pmatrix} x + 0.33 \\
		\layer_2(x) &= \relu\left(\begin{pmatrix} \phantom{-}0.78 & \phantom{-}2.17 & \phantom{-}0.39 \\ -0.72 & -0.75 & \phantom{-}1.02 \\ -0.14 & -0.64 & -0.32 \end{pmatrix} x + \begin{pmatrix} -2.42 \\ -1.23 \\ 0 \end{pmatrix}\right) &
		\NN &= \layer_3 \circ \layer_2 \circ \layer_1
	\end{align*}
	\caption{A DNN $\NN: \R^1 \to \R^1$ for approximating the parabola $f(x) = x^2/20$.}
	\label{fig:nn_parabola2d}
\end{figure}

We can use the preimage computation to learn about the function-approximation capabilities of a DNN.
As a case study, we look at the parabola $f(x) = x^2/20$.
We trained a DNN with two hidden layers, each with $3$ neurons, based on $100$ samples from the domain $[-20, 20]$.
The resulting DNN is given in \figref{nn_parabola2d}.

\begin{figure}[t]
	\centering
	\includegraphics[width=\textwidth,keepaspectratio]{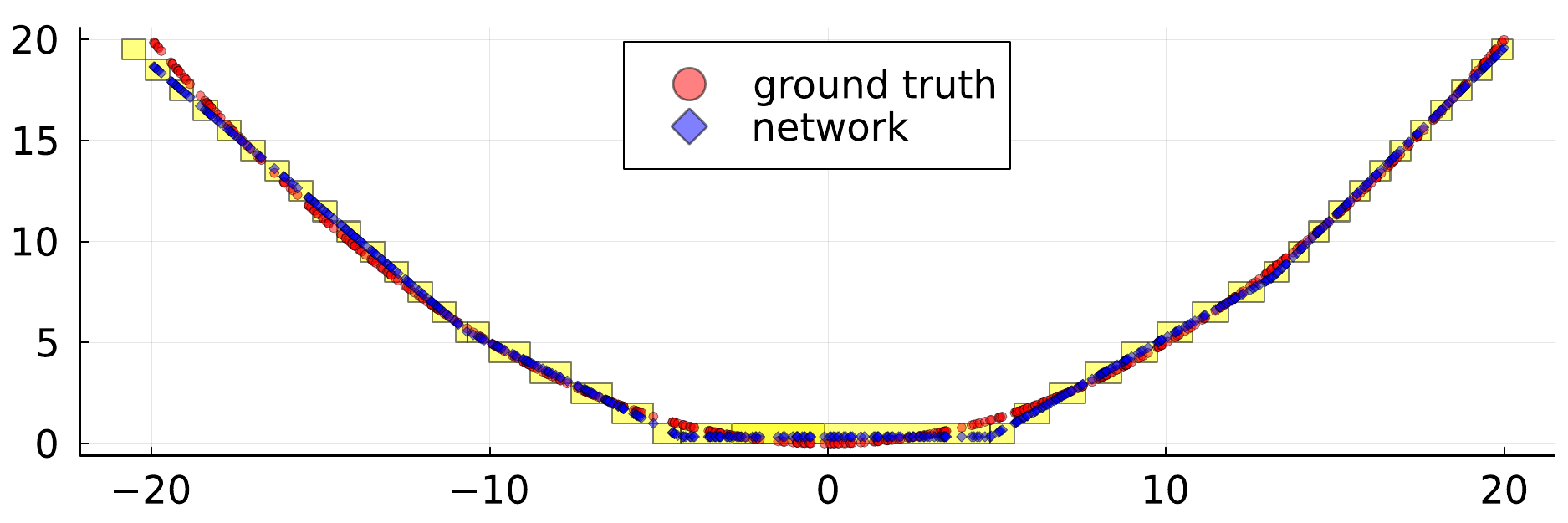}
	\caption{Preimage computation of the DNN in \figref{nn_parabola2d}.}
	\label{fig:parabola2d}
\end{figure}

In \figref{parabola2d} we plot the output of $500$ samples of the function $f$ (red) as well as the DNN approximation (blue).
In addition, we partition the codomain $[0, 20]$ into $20$ uniform intervals and compute the preimages (yellow boxes).
By construction, the blue samples lie inside the preimages.

In this case we could have obtained similar results via forward-image computation because, by assuming that the DNN approximates the training dataset well, we could have just partitioned the domain $[-20, 20]$ instead.
However, this does generally not work if we want to find the preimage of a subset in the codomain that the DNN does not map training data to, since we are clueless where in the domain to search.
For example, we can ask for the preimage in the interval $[100, 105]$.
This lets us study the generalization capabilities of the DNN.
The result is a union of two intervals, $[-80.88, -77.35]$ and $[68.46, 71.48]$.
The ground truth consists of the two intervals $\pm[44.72, 45.83]$.
We thus see that the DNN does not generalize well, and neither does it preserve the symmetry well.
However, the DNN seems to preserve both a negative and a positive preimage.

We can also easily prove that the DNN does not output negative numbers.
For that we compute $\NN^{-1}(\{y : y \leq 0\}) = \emptyset$.

\subsection{Approximation schemes}

The bottleneck of the calculations is the inverse activation function, due to the partitioning.
This motivates to seek approximate solutions.
We are generally interested in solutions that either contain the true solution (overapproximation) or are contained in the true solution (underapproximation).

\smallskip

A simple approach to compute an underapproximation of the preimage considers the partitioning of the sets (as, e.g., in \figref{ex_backward_l3relu}) as a search space and selects only one of the sets to continue with.
In general, this search may end up in an empty set (dead end), in which case one has to backtrack and pick the next set using some search heuristics (e.g., breadth-first or depth-first search).
In experiments on DNNs with many neurons per layer, where explicit partitioning is infeasible, we noticed that most of the sets are indeed empty.

\smallskip

Instead of underapproximations, we can also consider overapproximations.
Using abstract interpretation, we can choose an abstract domain to simplify the calculations.
Here we consider the standard interval approximation.

\begin{figure}[t]
	\centering
	\includegraphics[width=\textwidth,keepaspectratio]{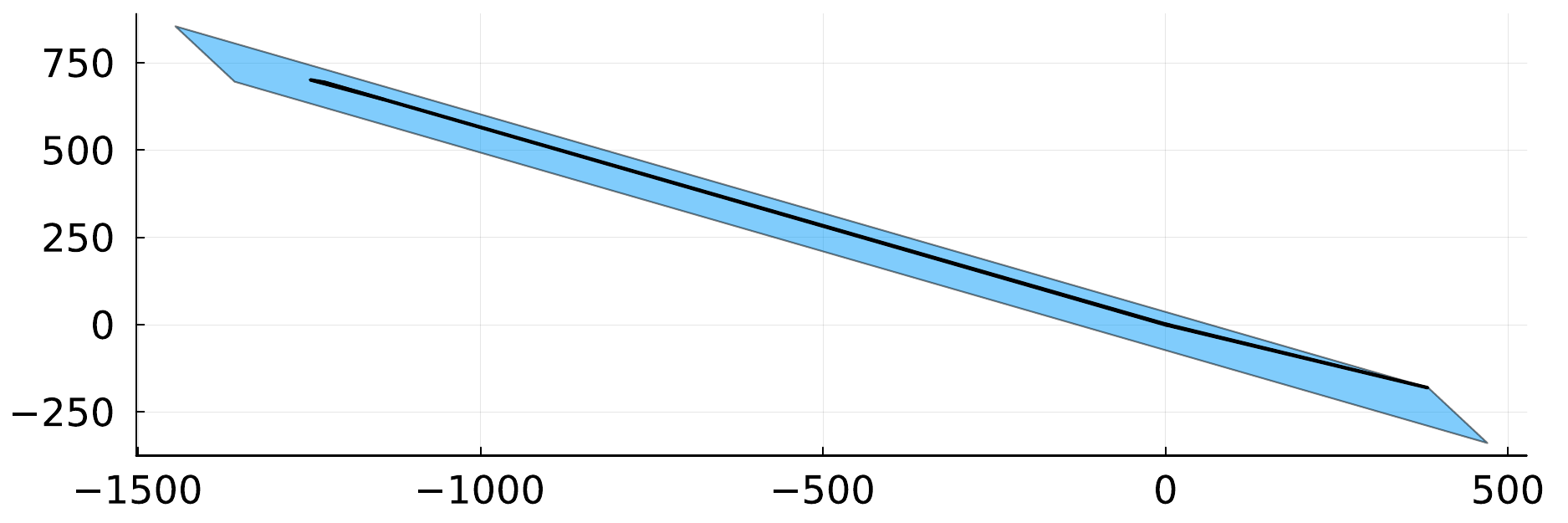}
	\caption{Preimage computation of the DNN in \figref{nn_example} but with leaky ReLUs ($\lrelu{0.01}$ and $\lrelu{0.02}$).
	The thick black shape is the exact preimage.
	The blue shape is the preimage where we applied a box approximation inverting the activations.}
	\label{fig:leaky_relu}
\end{figure}

As noted before, in general, the preimage of a DNN is unbounded, which makes an interval approximation useless.
To make sure that we receive a bounded preimage for each layer, we consider injective activation functions and layers of the same size.
We use the DNN in \figref{nn_example} but with leaky ReLUs ($\lrelu{0.01}$ in layer~$1$ and $\lrelu{0.02}$ in layer~$2$).
First we compute an overapproximation $\Y$ of the image of the domain (i.e., $\Y \supseteq \NN([0, 1]^2)$) using interval approximation.
Then we compute the preimage $\NN(\Y)$, using the exact algorithm, as well as obtain an overapproximation by applying an interval approximation for inverting the leaky-ReLU activations.
The approximate calculations are ${\approx}\,100$x faster than with the exact algorithm, but they also yield a coarser result, as shown in \figref{leaky_relu}.

Interval approximation is also attractive for activations that are not piecewise affine.
We discuss this in the next experiment.

\subsection{Forward-backward computation}

The approach in \cite{Thrun94} propagates intervals forward and backward in a DNN, just as the interval approximation explained above.
The benefit of this scheme is that it applies to activations that are not piecewise affine, such as the sigmoid function $\sigmoid(x) = 1 / (1 + e^{-x})$.
Backward propagation of intervals is easy for strictly monotonic activations like the sigmoid: one just applies the inverse function to the lower and upper bound.

\begin{figure}[t]
	\centering
	\begin{align*}
		\layer_1(x) &= \sigmoid\left(\begin{pmatrix} -4.60248 & 4.74295 \\ -3.19378 & 2.90011\end{pmatrix} x + \begin{pmatrix} \phantom{-}2.74108 \\ -1.49695 \end{pmatrix}\right) &
		\NN &= \layer_2 \circ \layer_1 \\
		\layer_2(x) &= \sigmoid\left(\begin{pmatrix} -4.57199 & 4.64925\end{pmatrix} x + 2.10176 \right)
	\end{align*}
	\caption{A DNN to approximate the XOR function (taken from~\cite{Thrun94}).}
	\label{fig:nn_thrun94}
\end{figure}

We repeat an experiment from~\cite{Thrun94}.
The DNN (given in \figref{nn_thrun94}) uses sigmoid activations and was trained to implement a real approximation of the XOR function, i.e., it should output a value close to $1$ if and only if one of the inputs is close to~$1$.
The input domain is $[0, 1]^2$ and we consider five scenarios where we add additional input or output constraints.

\begin{figure}[t]
	\centering
	input neuron $1$ \hspace*{.2mm} input neuron $2$ \hspace*{.2mm} hidden neuron $1$ \hspace*{.2mm} hidden neuron $2$ \hspace*{.2mm} output neuron
	
	\includegraphics[width=0.19\textwidth,keepaspectratio]{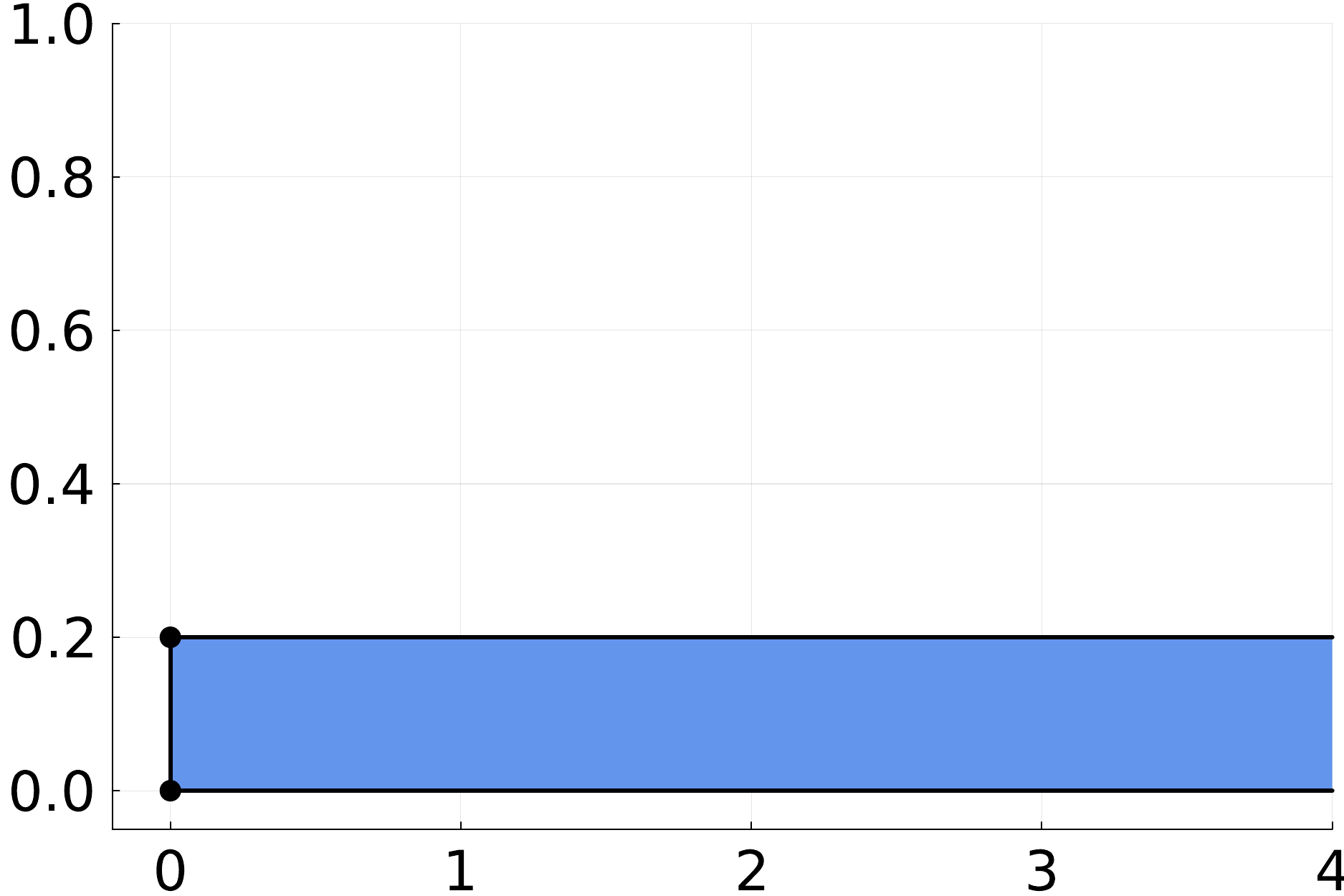}
	\includegraphics[width=0.19\textwidth,keepaspectratio]{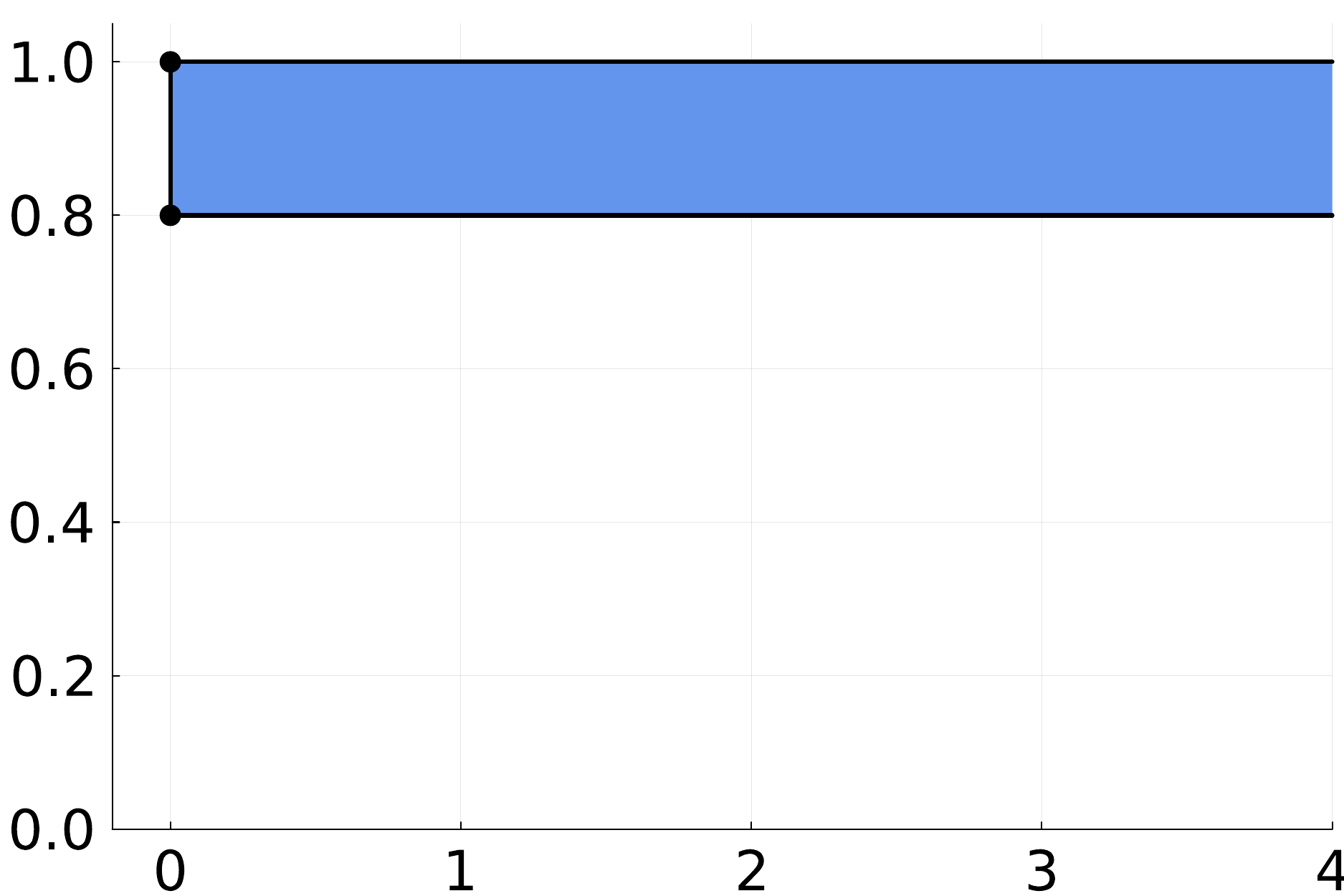}
	\includegraphics[width=0.19\textwidth,keepaspectratio]{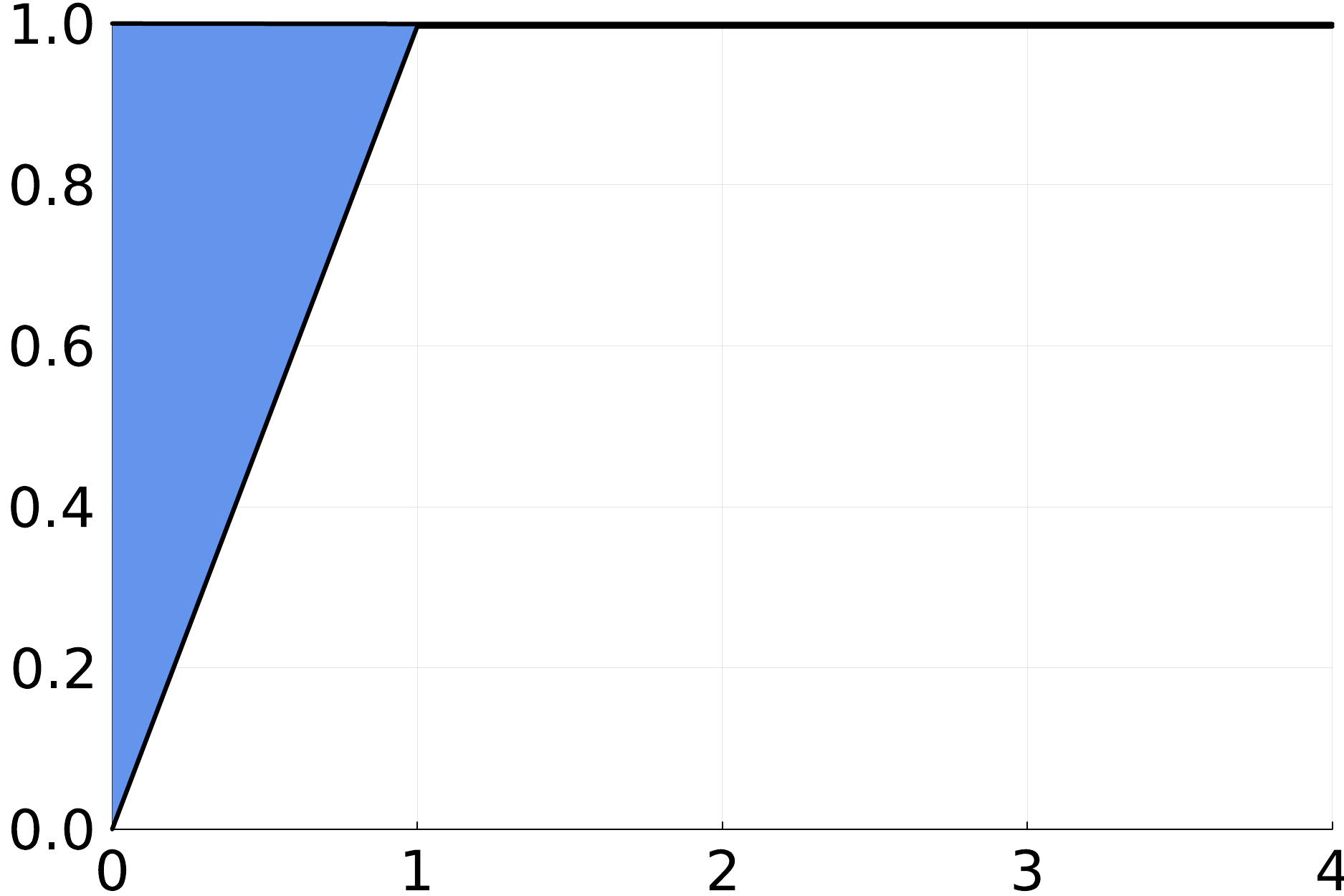}
	\includegraphics[width=0.19\textwidth,keepaspectratio]{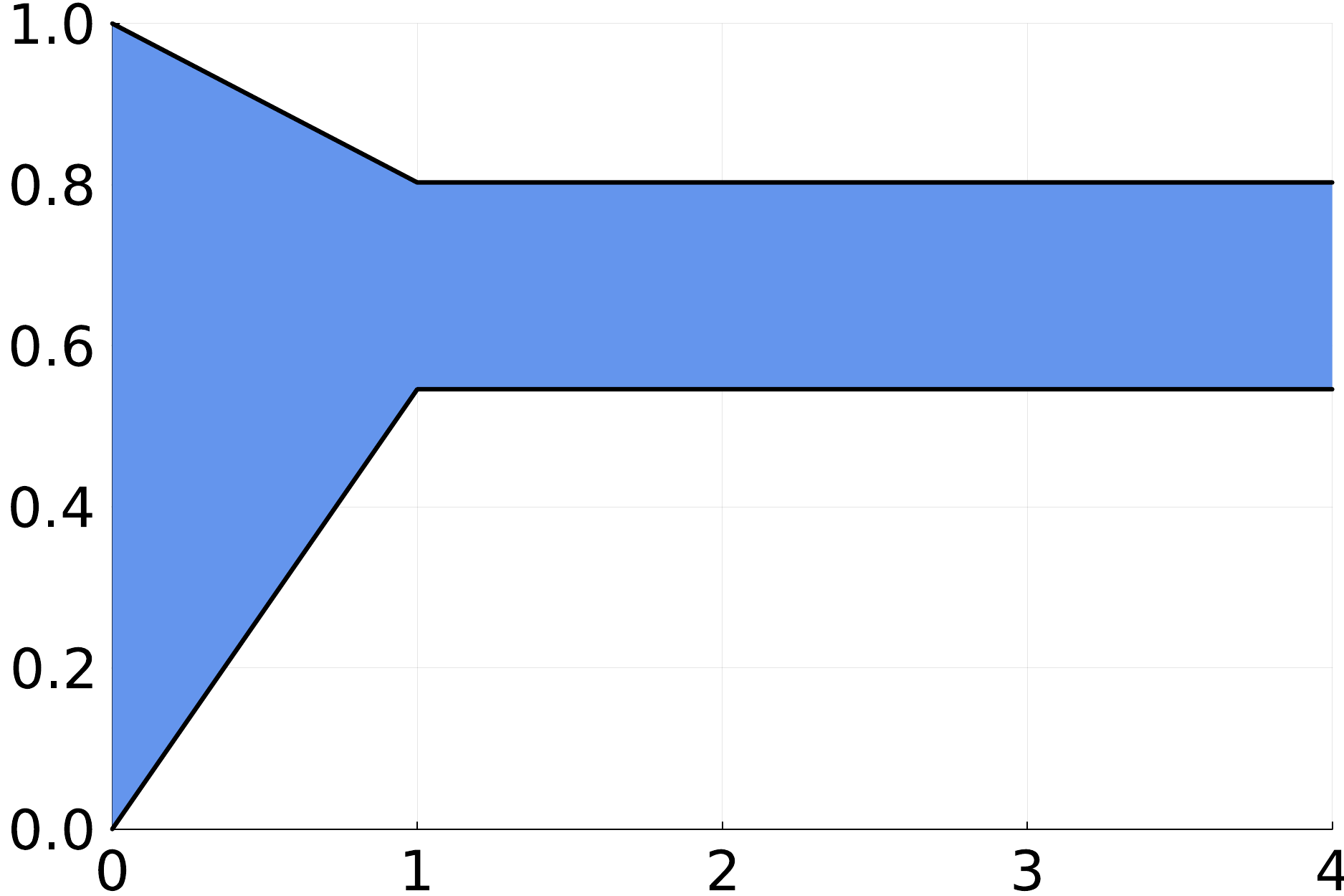}
	\includegraphics[width=0.19\textwidth,keepaspectratio]{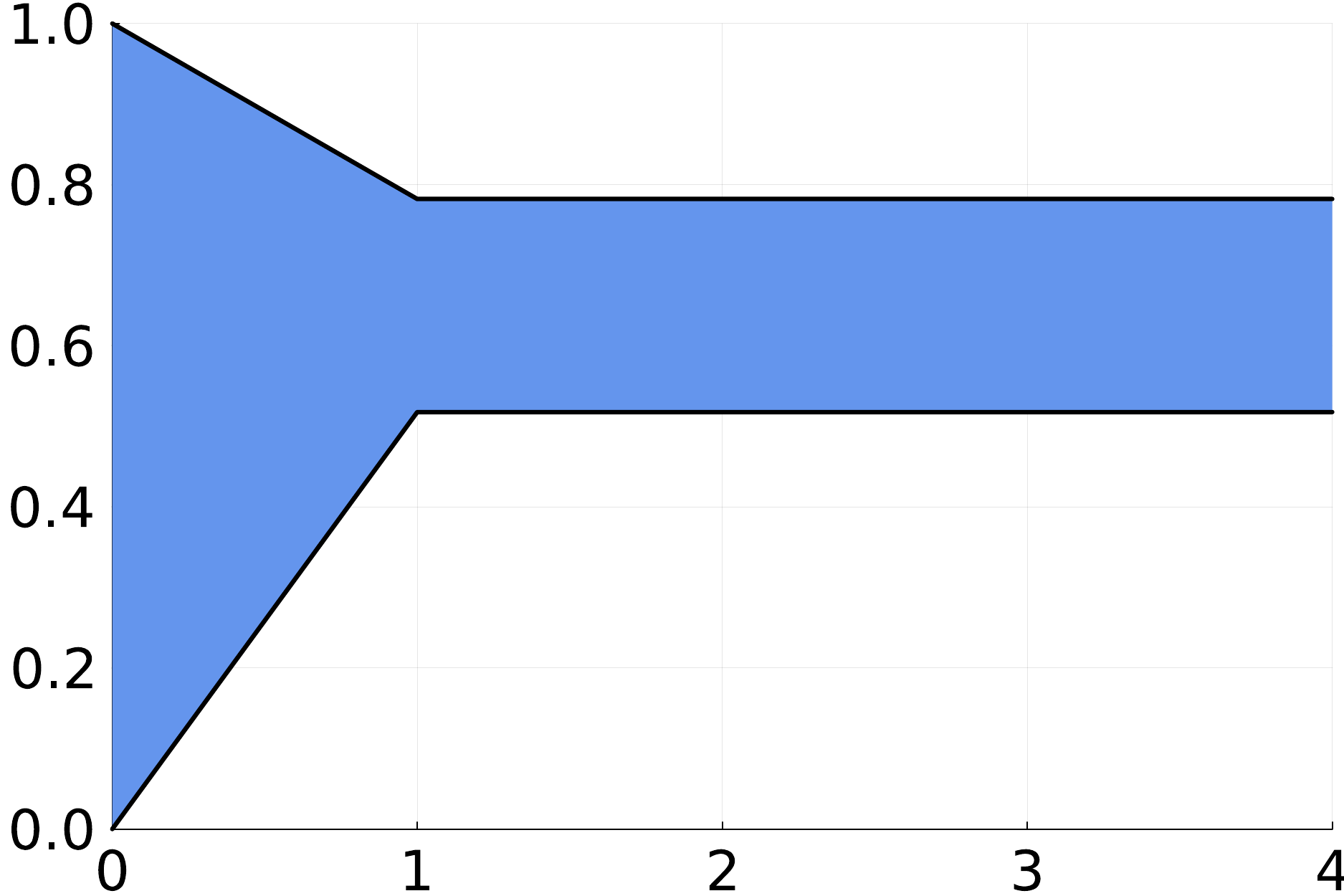}
	
	\includegraphics[width=0.19\textwidth,keepaspectratio]{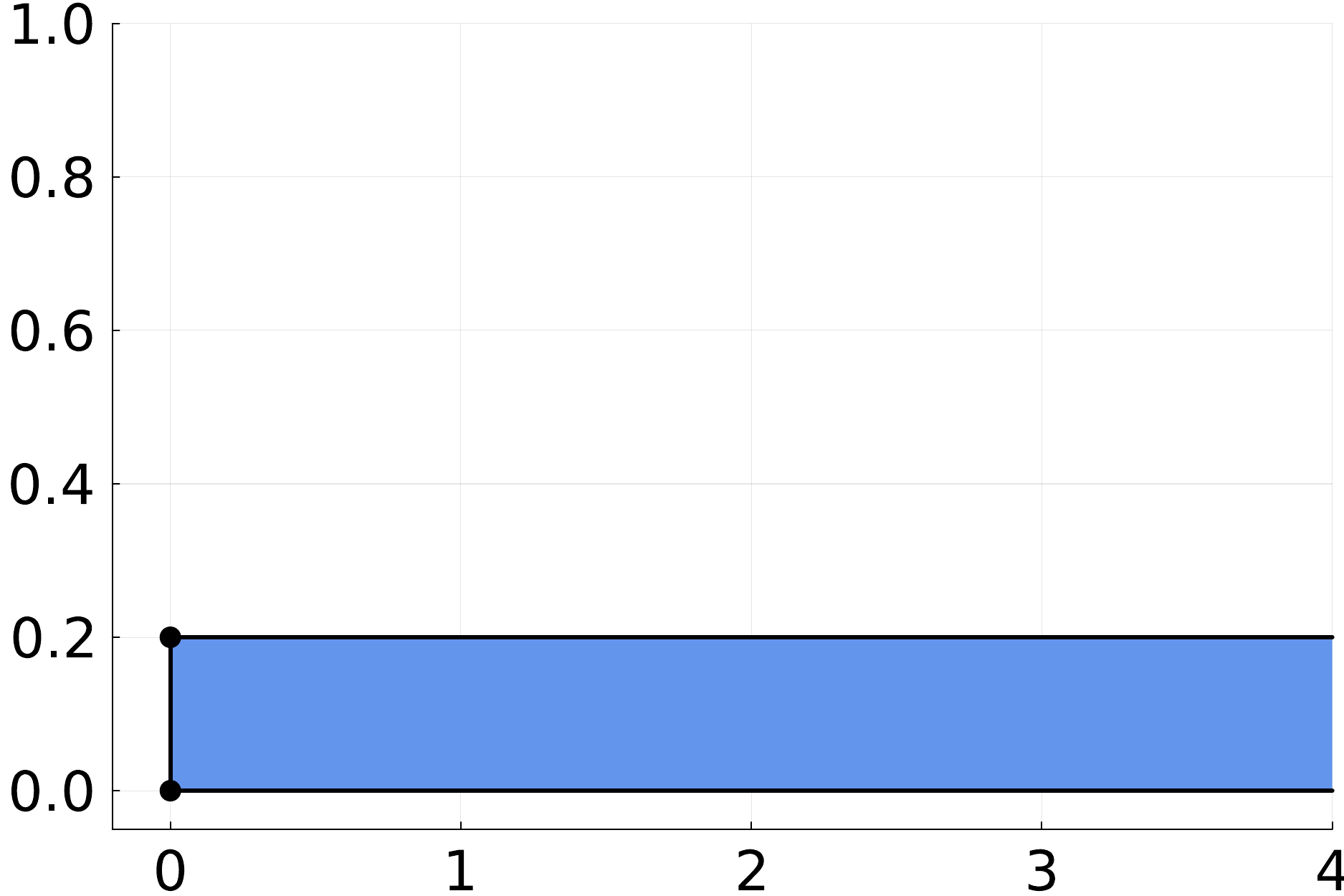}
	\includegraphics[width=0.19\textwidth,keepaspectratio]{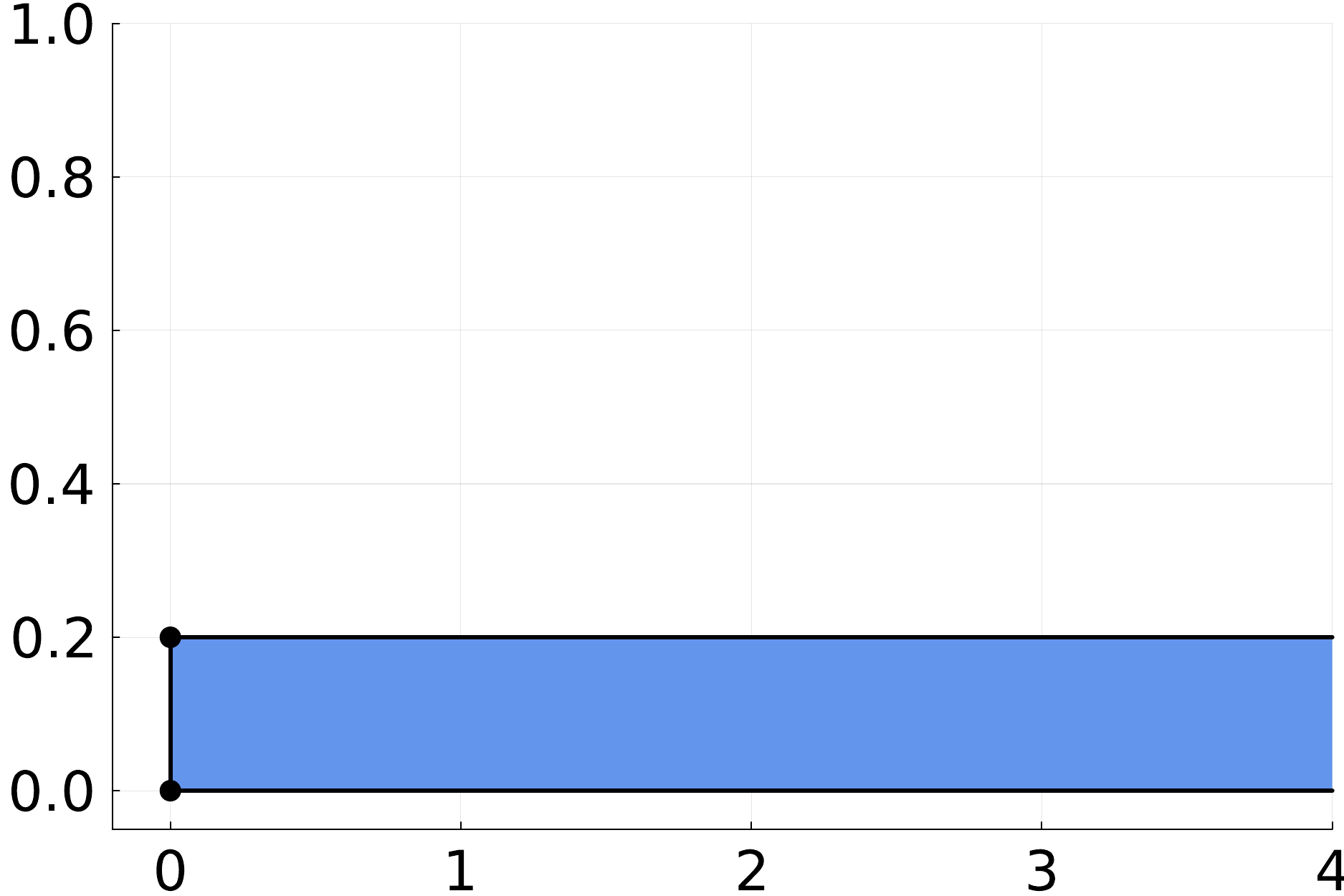}
	\includegraphics[width=0.19\textwidth,keepaspectratio]{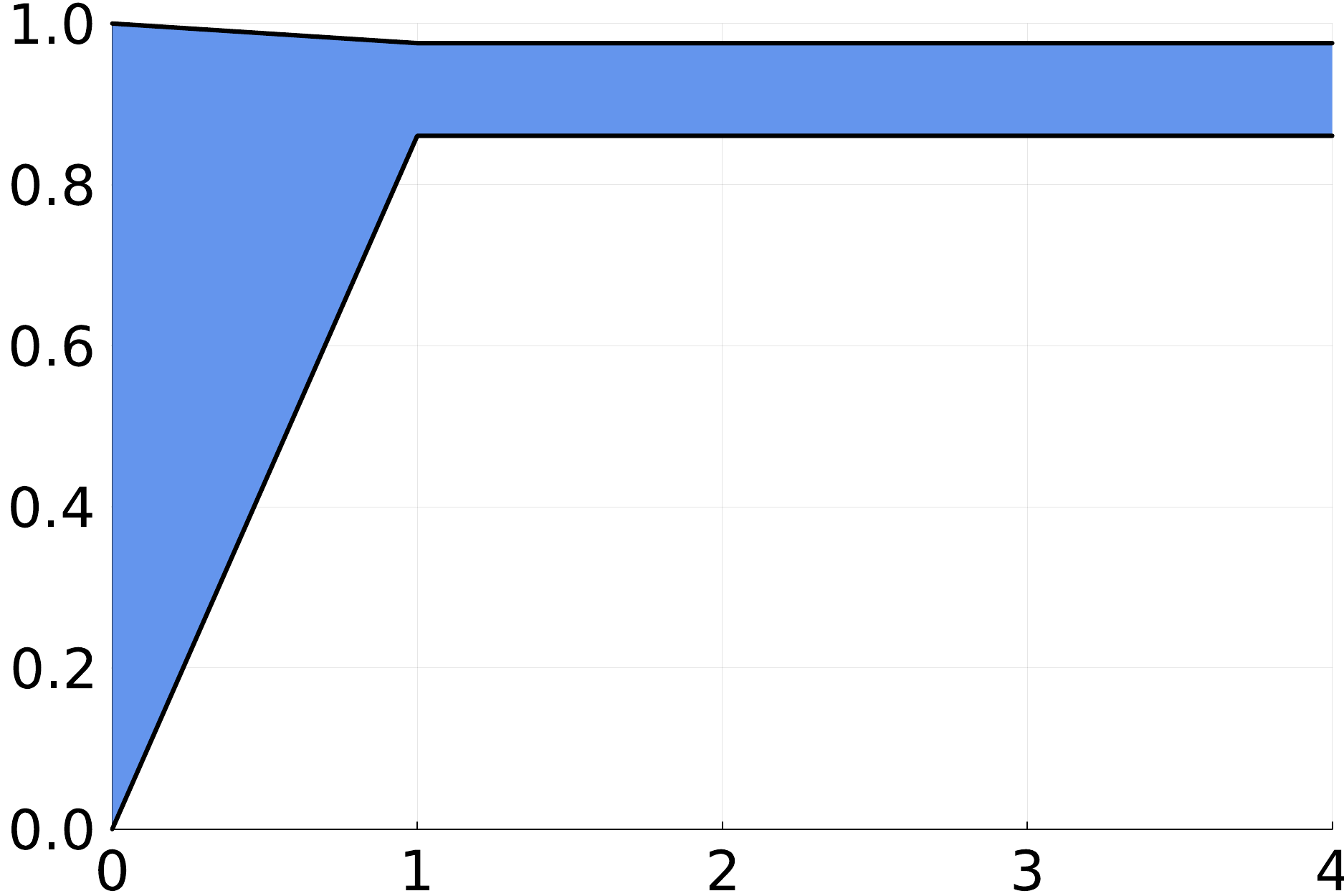}
	\includegraphics[width=0.19\textwidth,keepaspectratio]{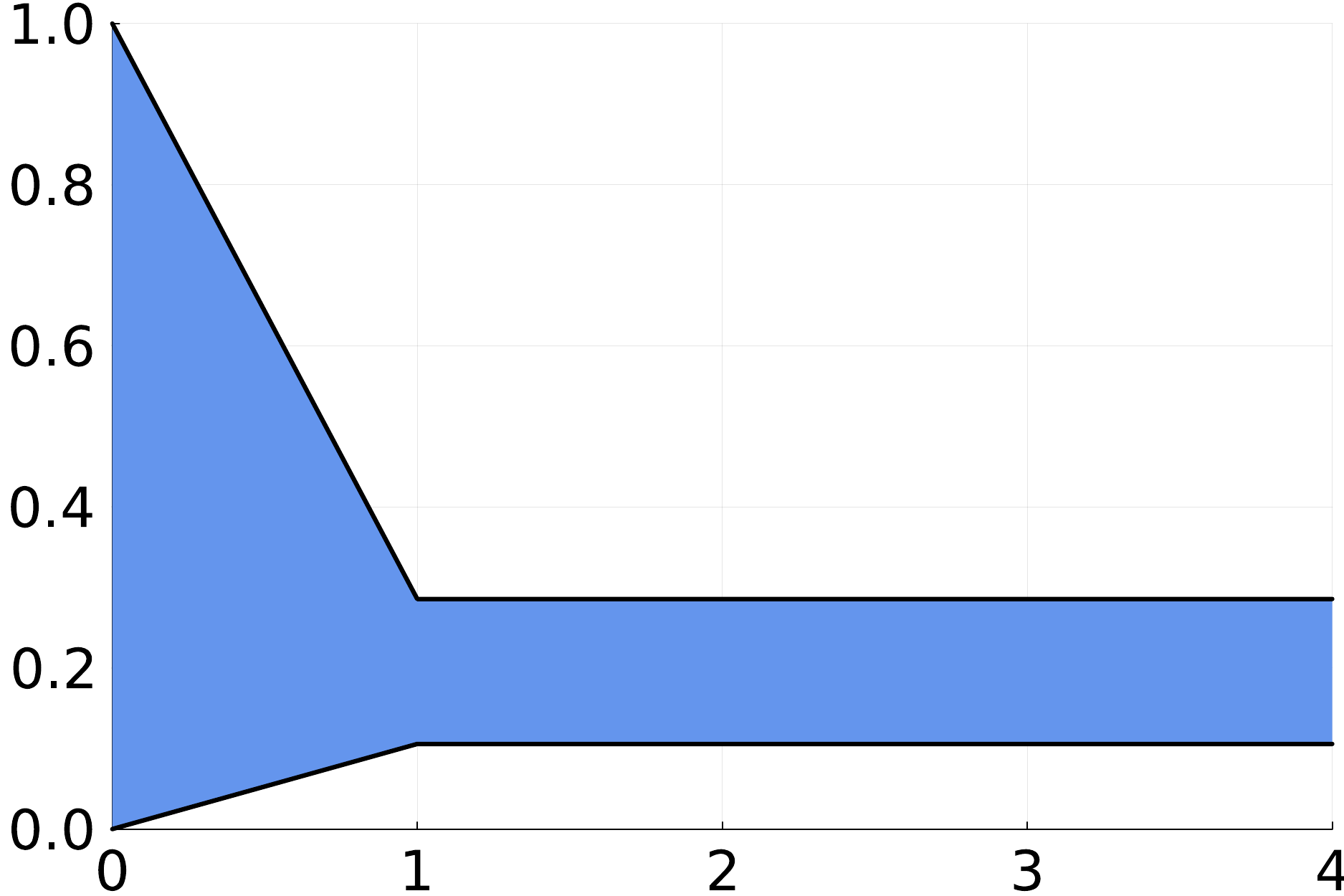}
	\includegraphics[width=0.19\textwidth,keepaspectratio]{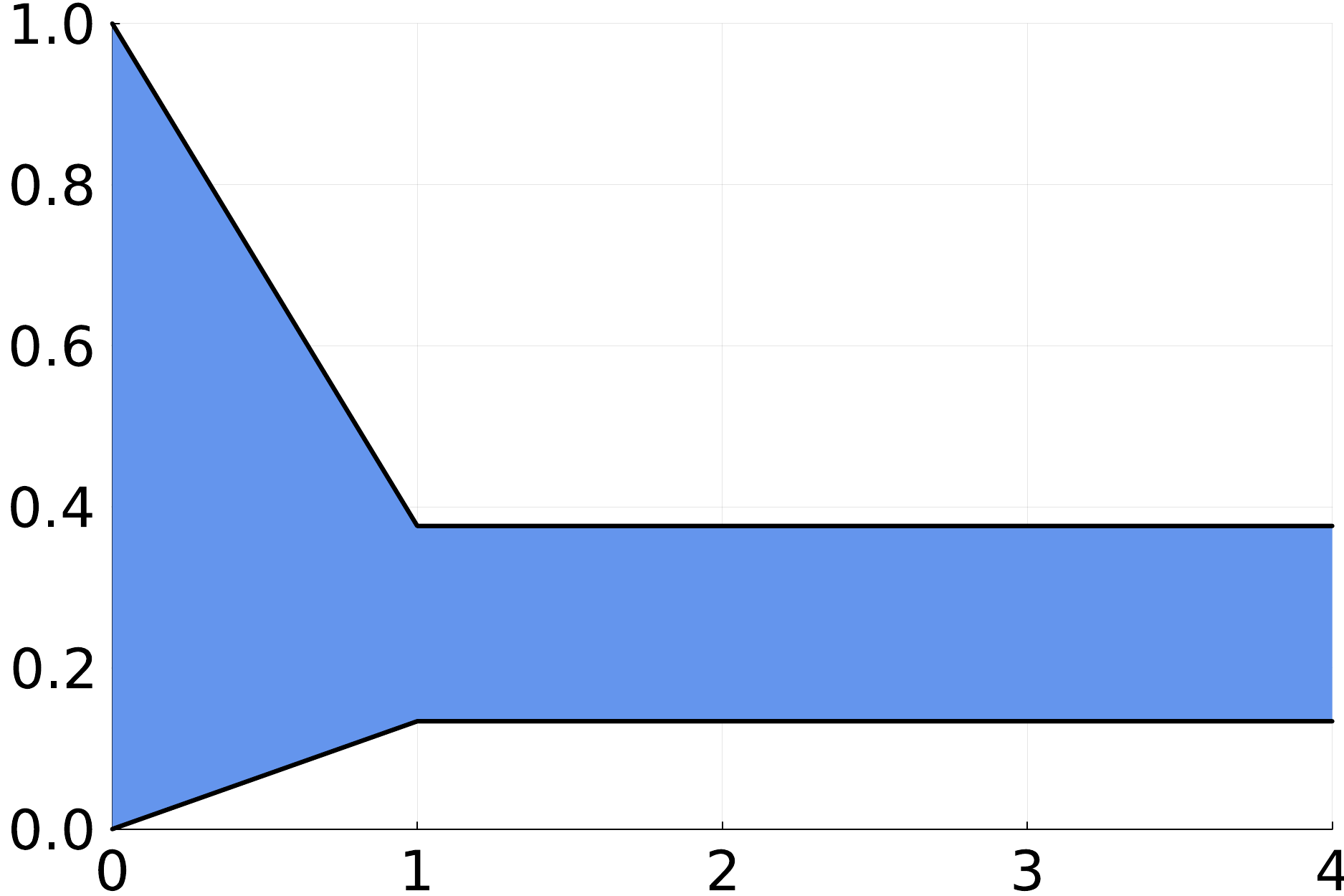}
	
	\includegraphics[width=0.19\textwidth,keepaspectratio]{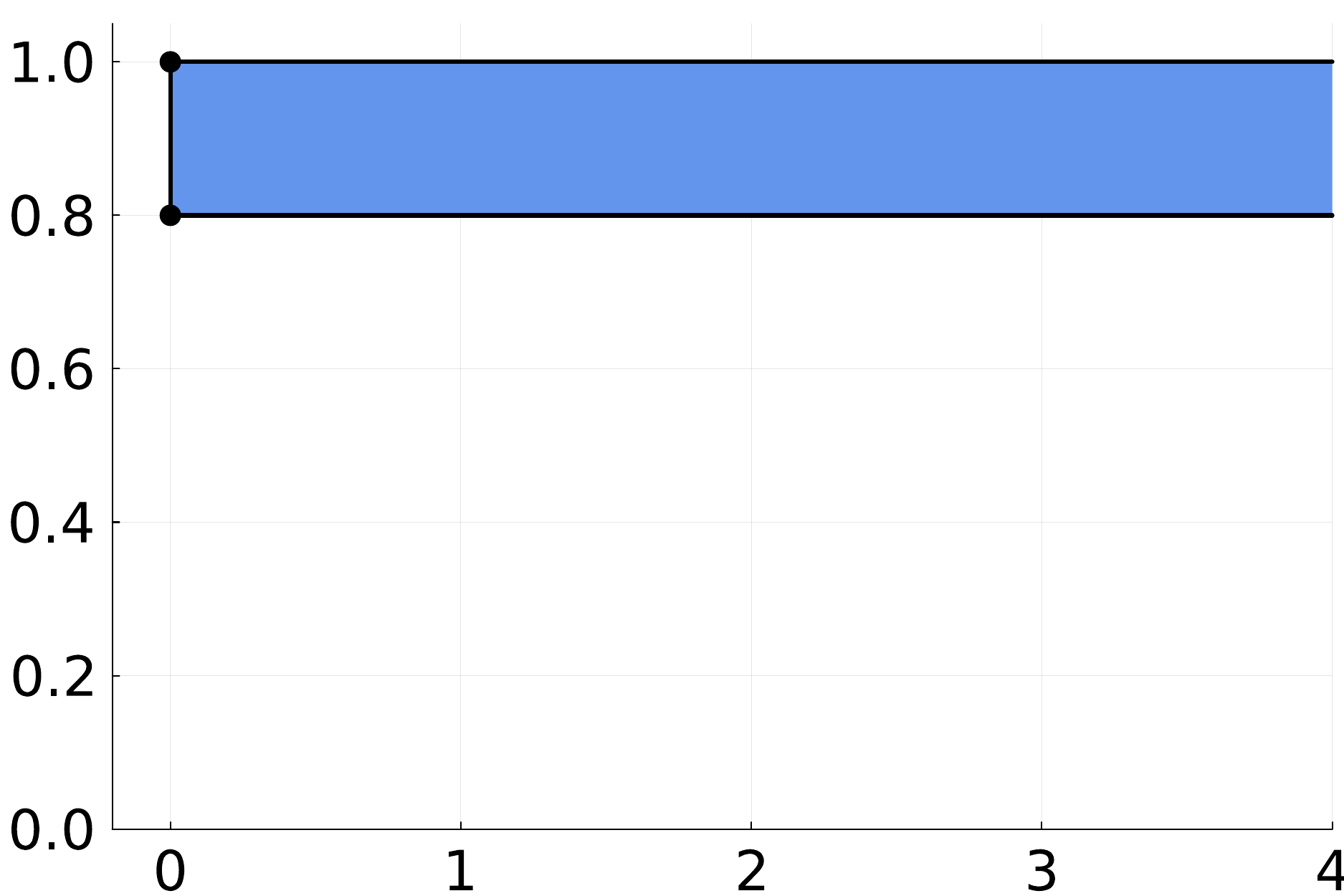}
	\includegraphics[width=0.19\textwidth,keepaspectratio]{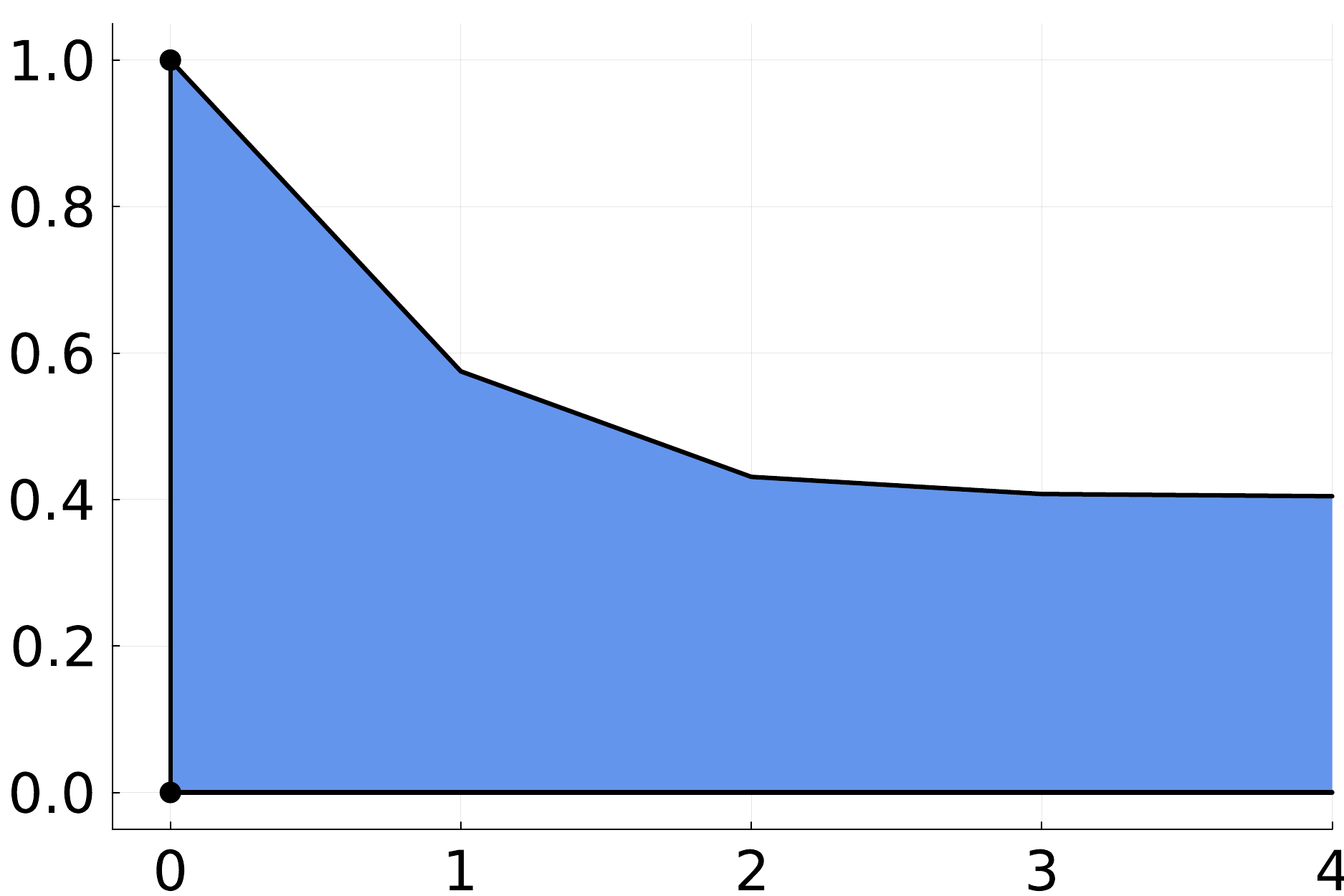}
	\includegraphics[width=0.19\textwidth,keepaspectratio]{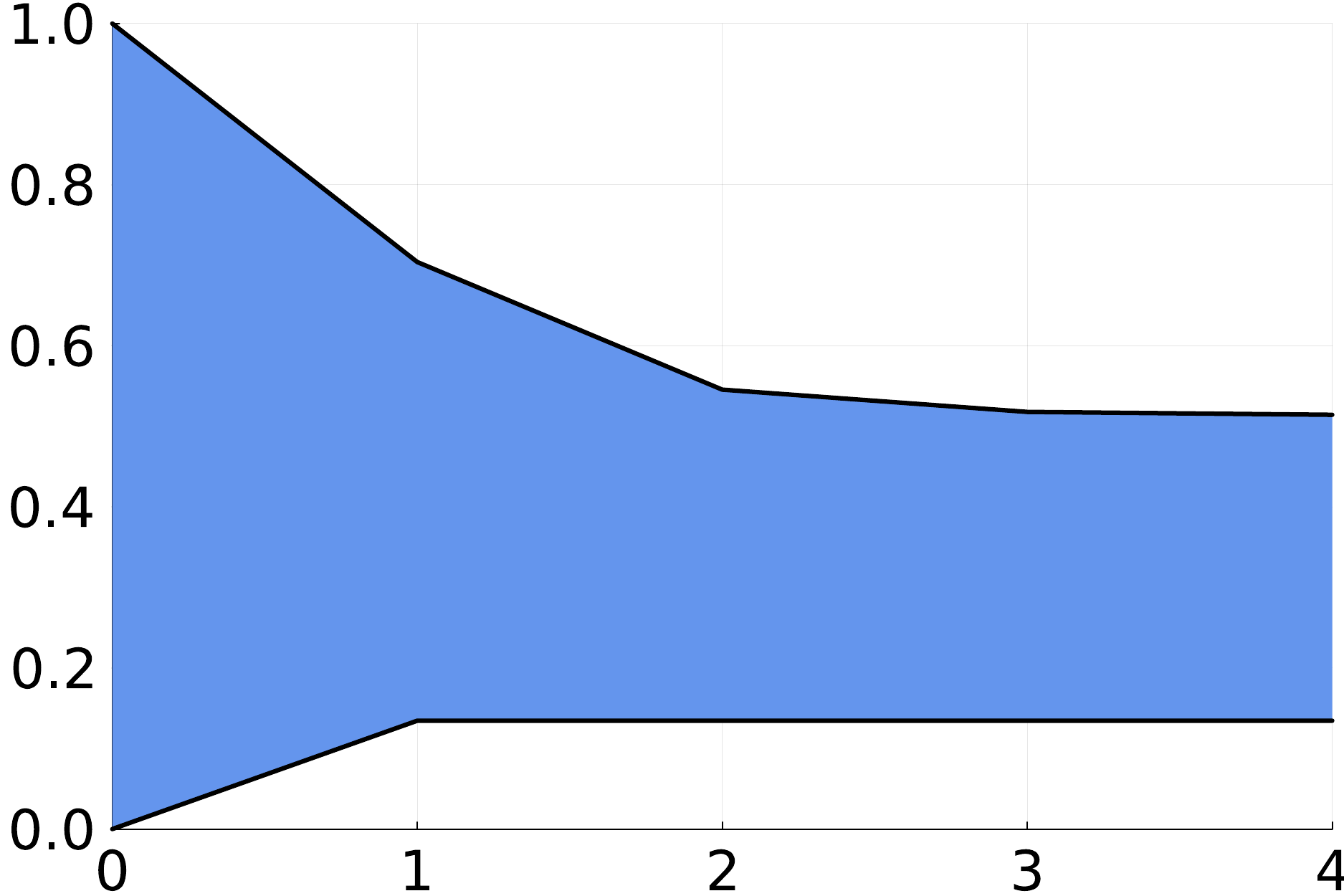}
	\includegraphics[width=0.19\textwidth,keepaspectratio]{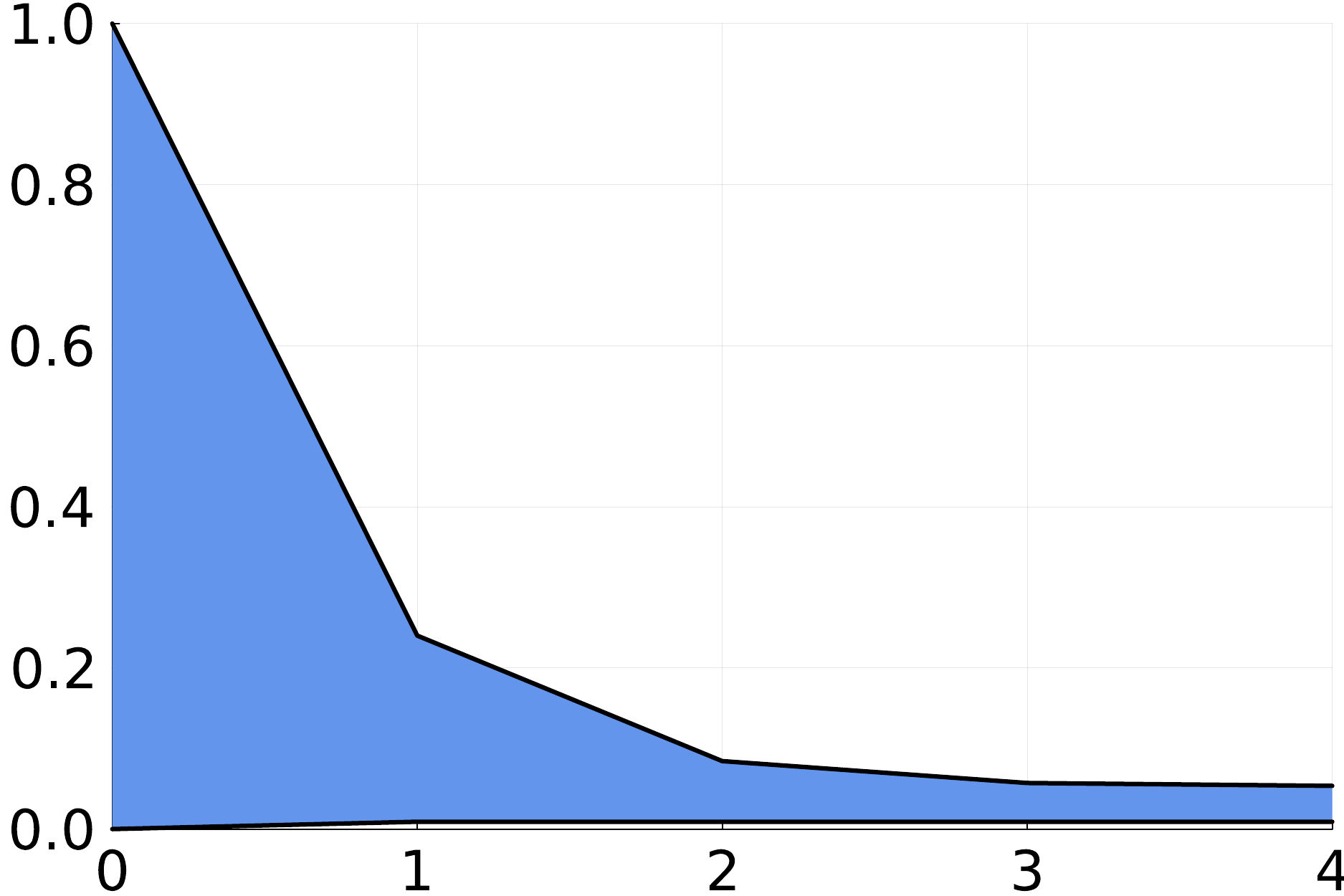}
	\includegraphics[width=0.19\textwidth,keepaspectratio]{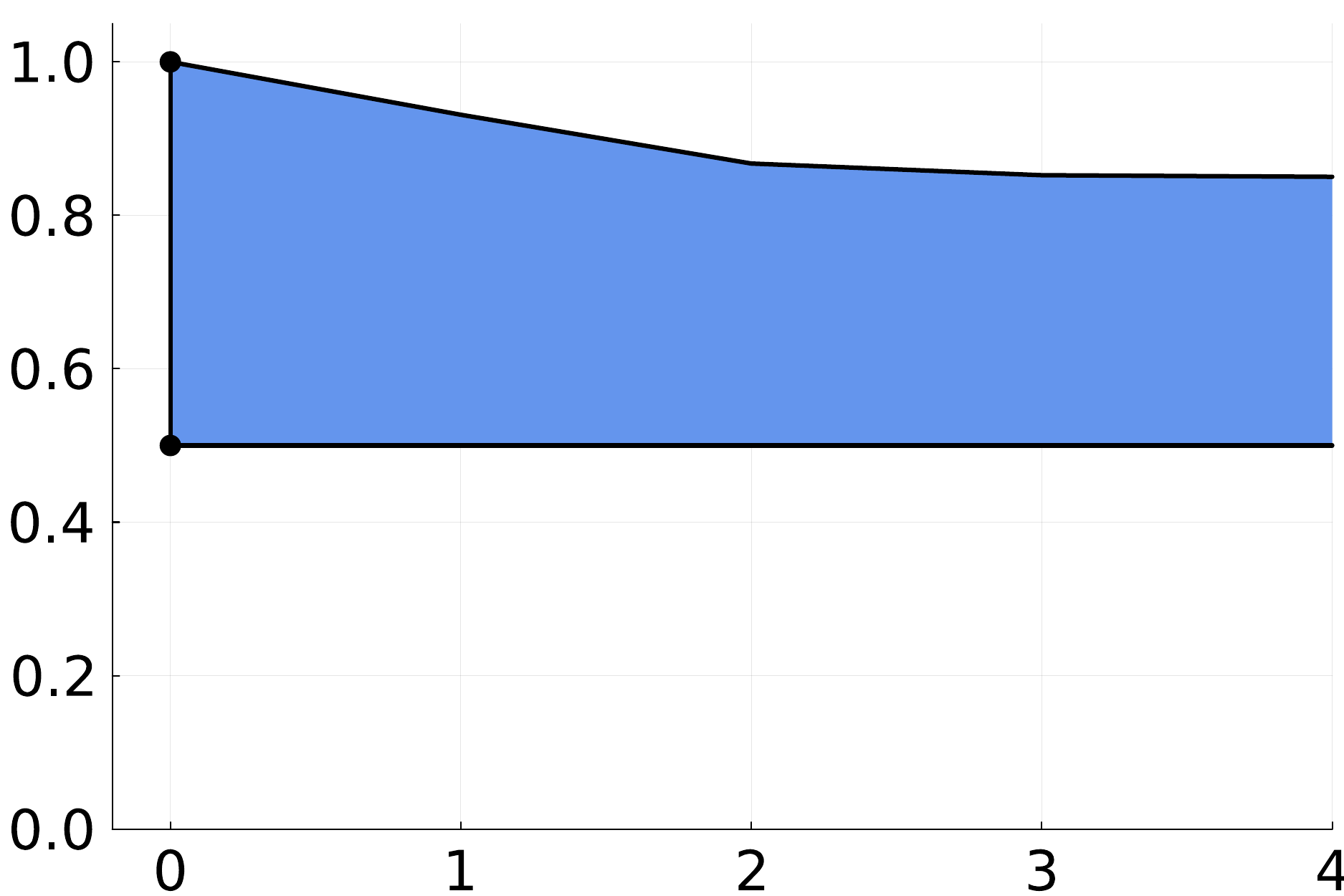}
	
	\includegraphics[width=0.19\textwidth,keepaspectratio]{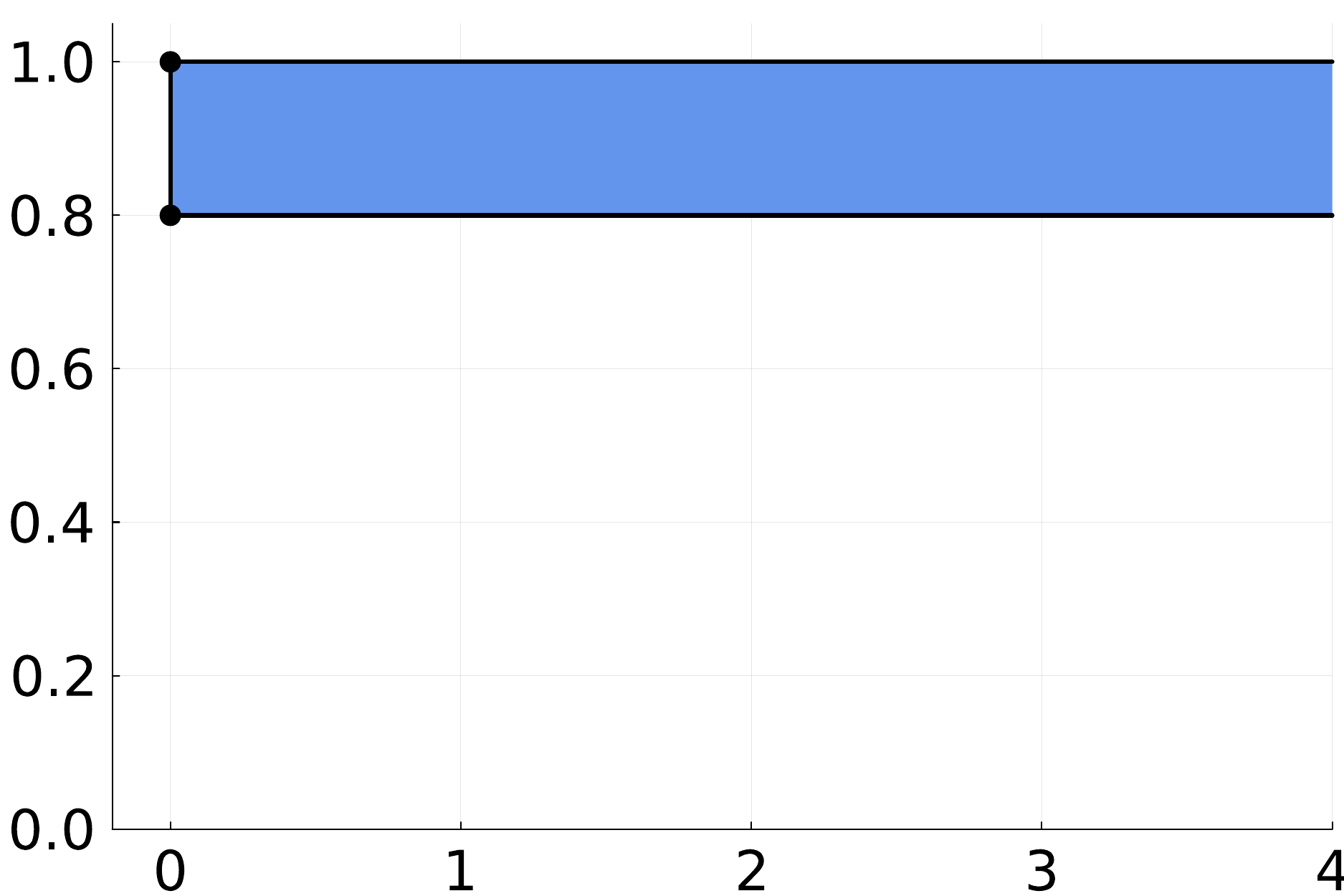}
	\includegraphics[width=0.19\textwidth,keepaspectratio]{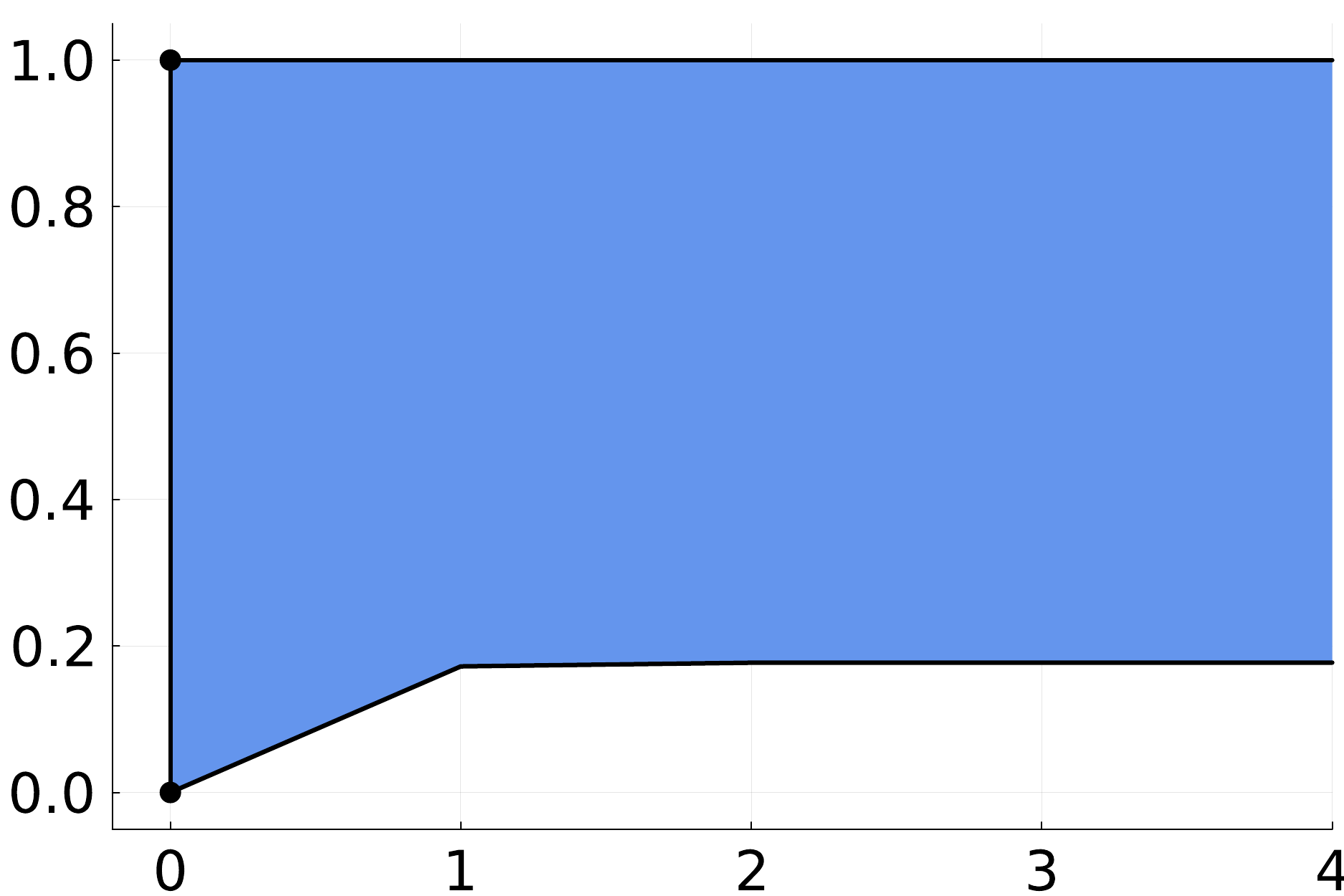}
	\includegraphics[width=0.19\textwidth,keepaspectratio]{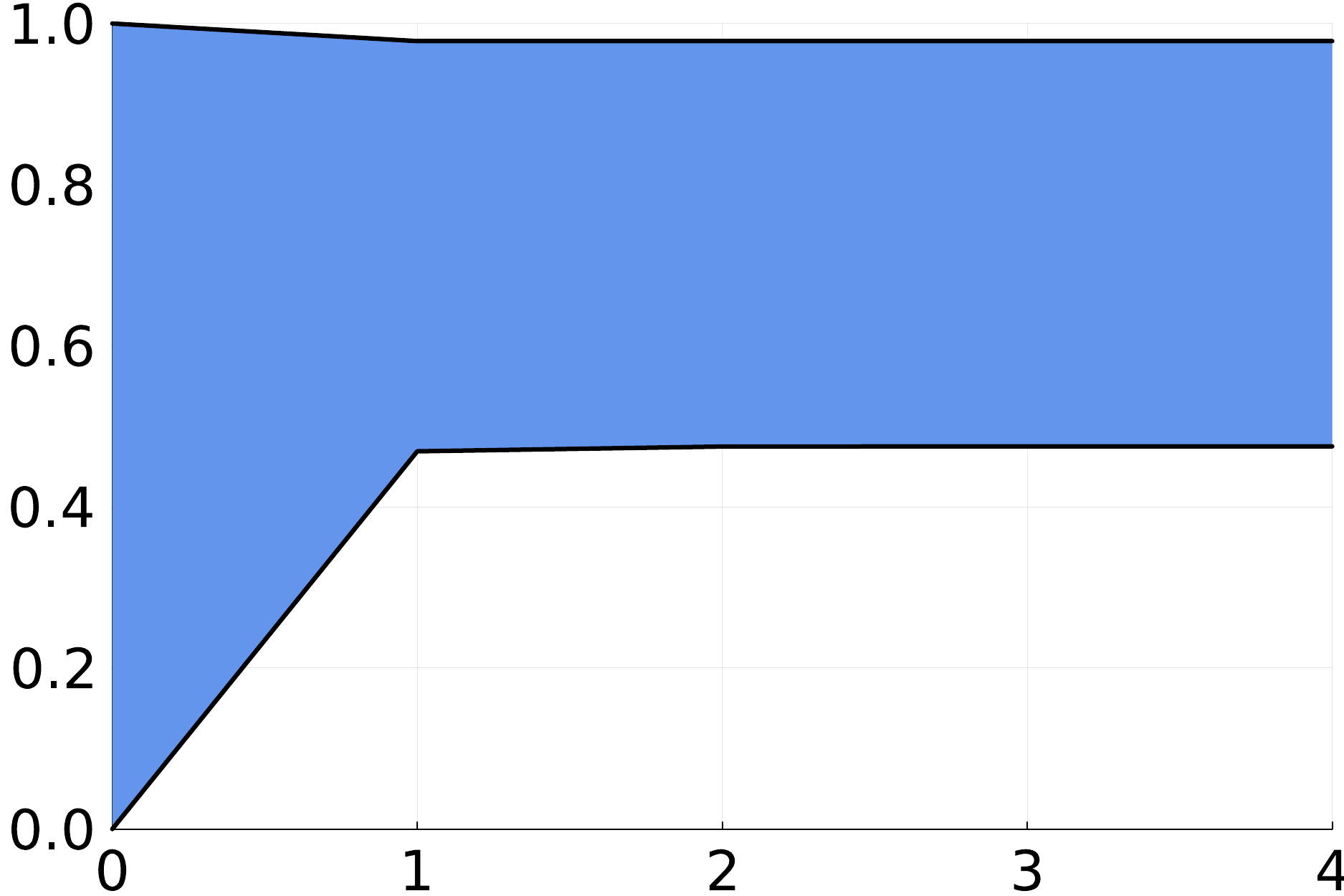}
	\includegraphics[width=0.19\textwidth,keepaspectratio]{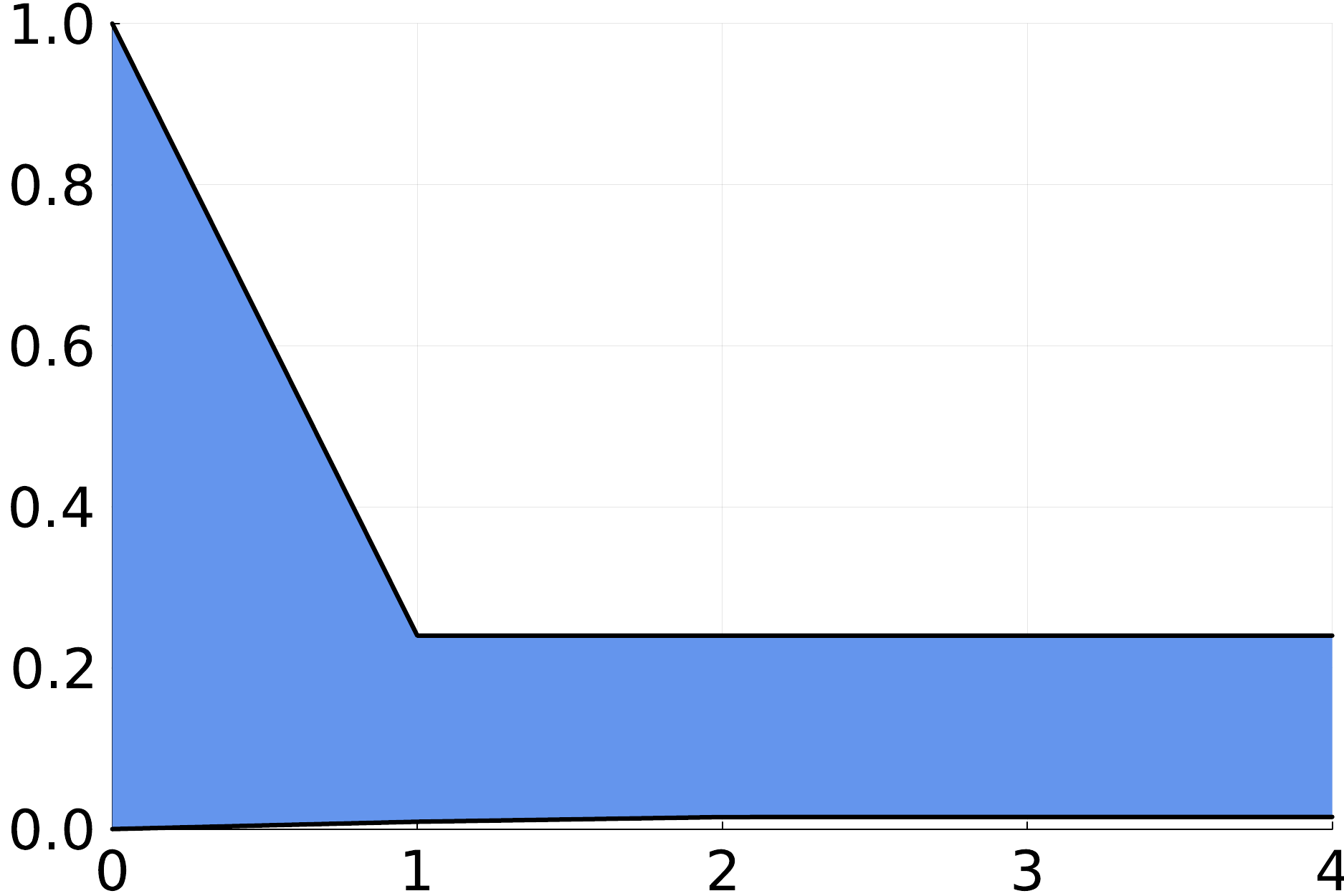}
	\includegraphics[width=0.19\textwidth,keepaspectratio]{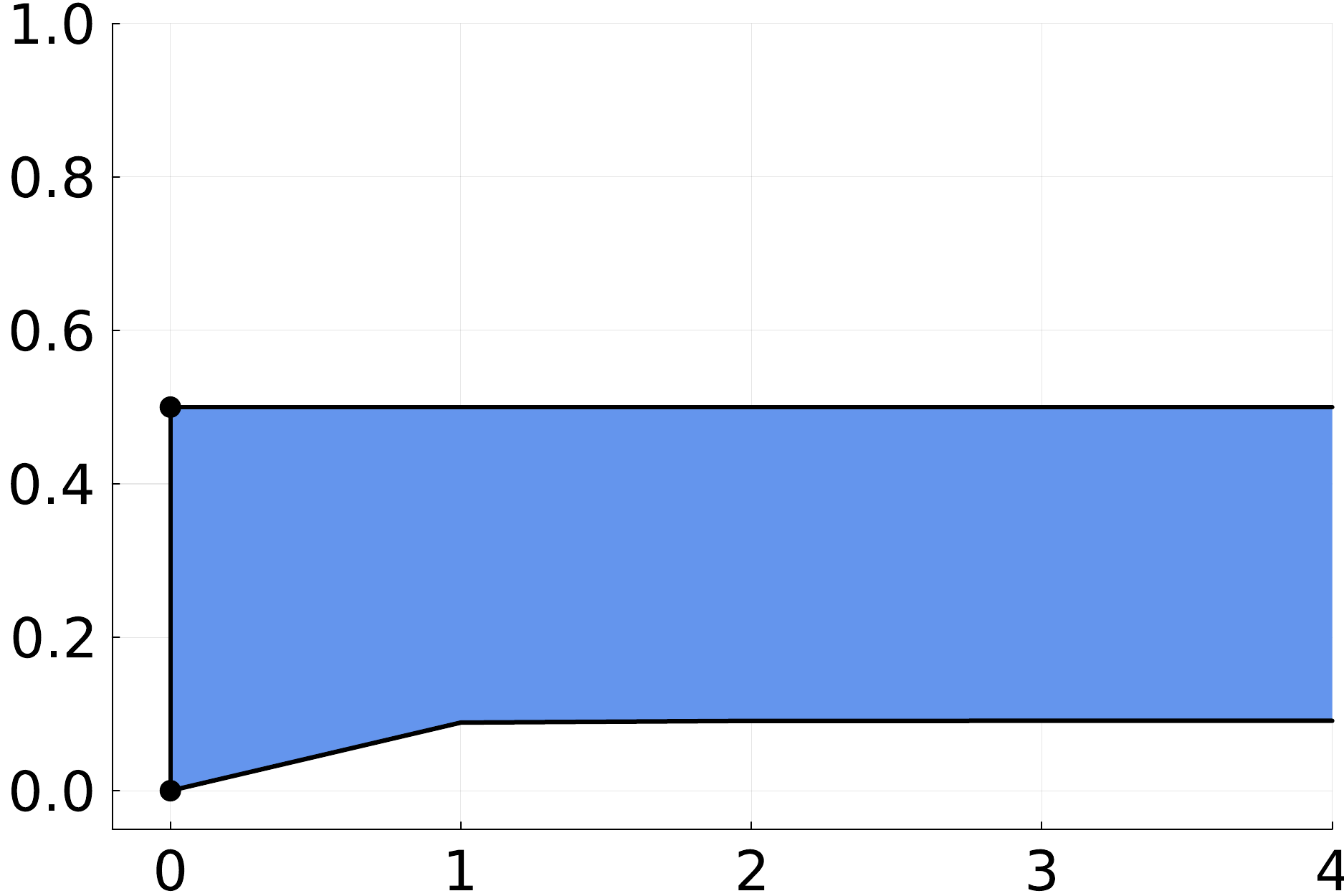}
	
	\includegraphics[width=0.19\textwidth,keepaspectratio]{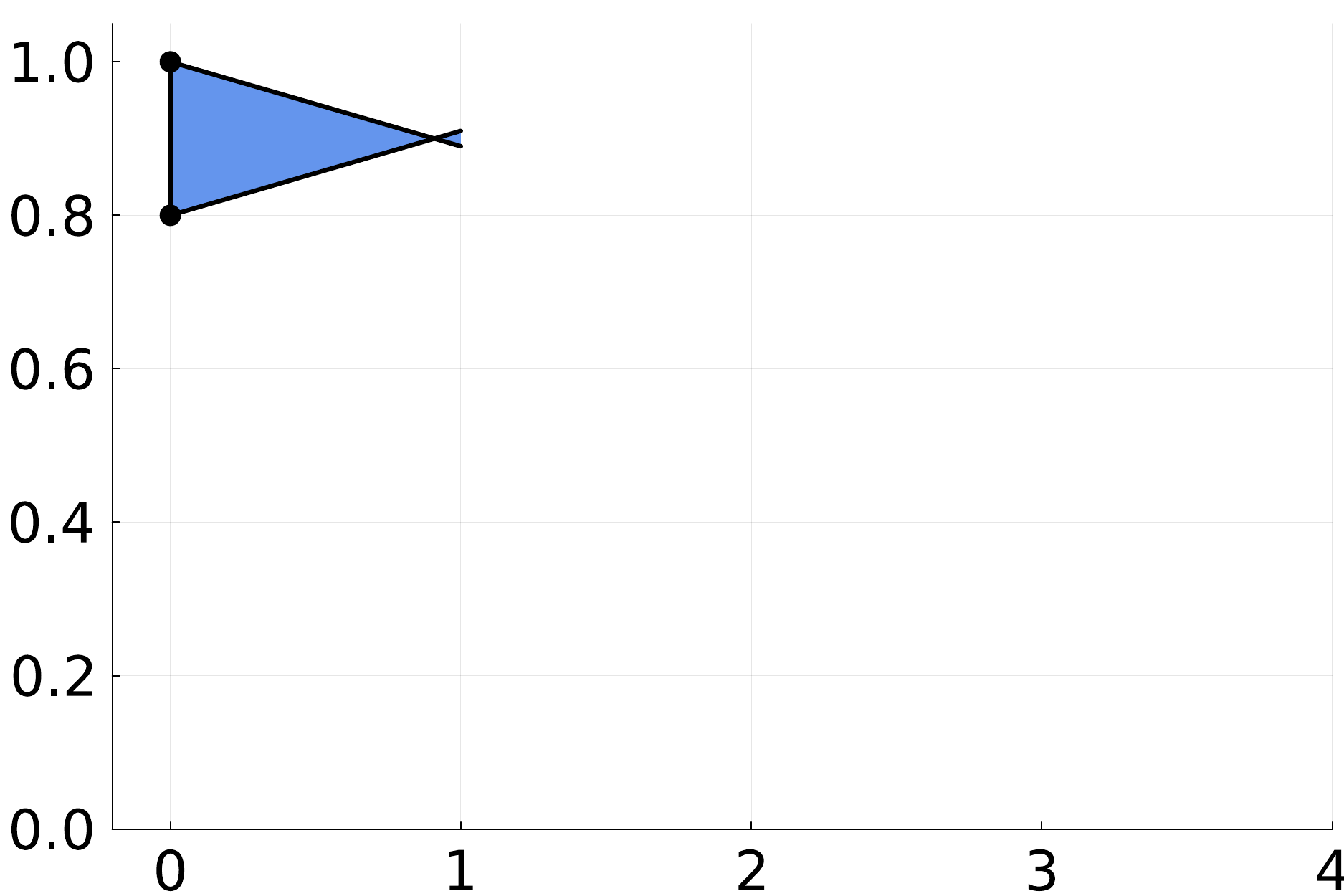}
	\includegraphics[width=0.19\textwidth,keepaspectratio]{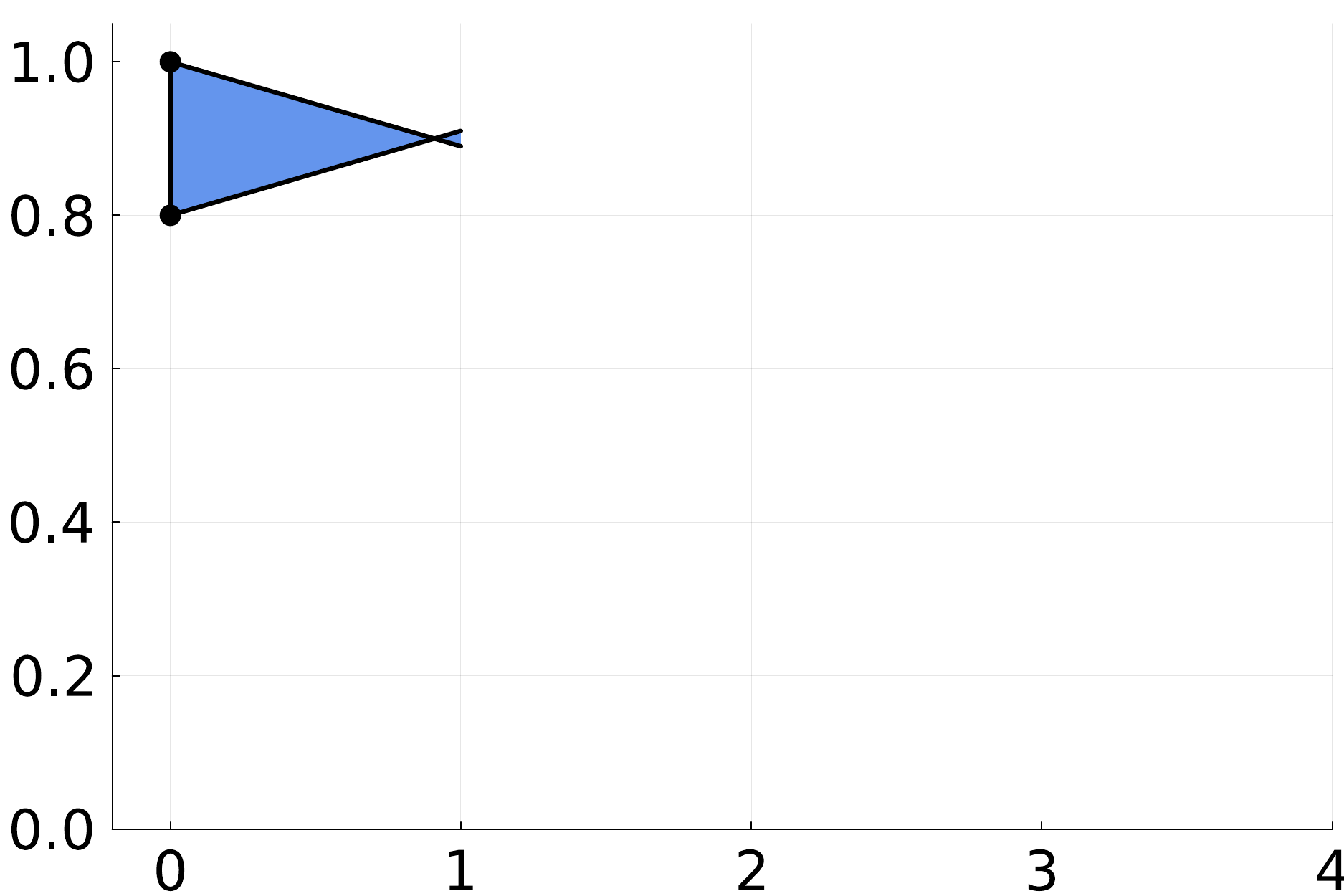}
	\includegraphics[width=0.19\textwidth,keepaspectratio]{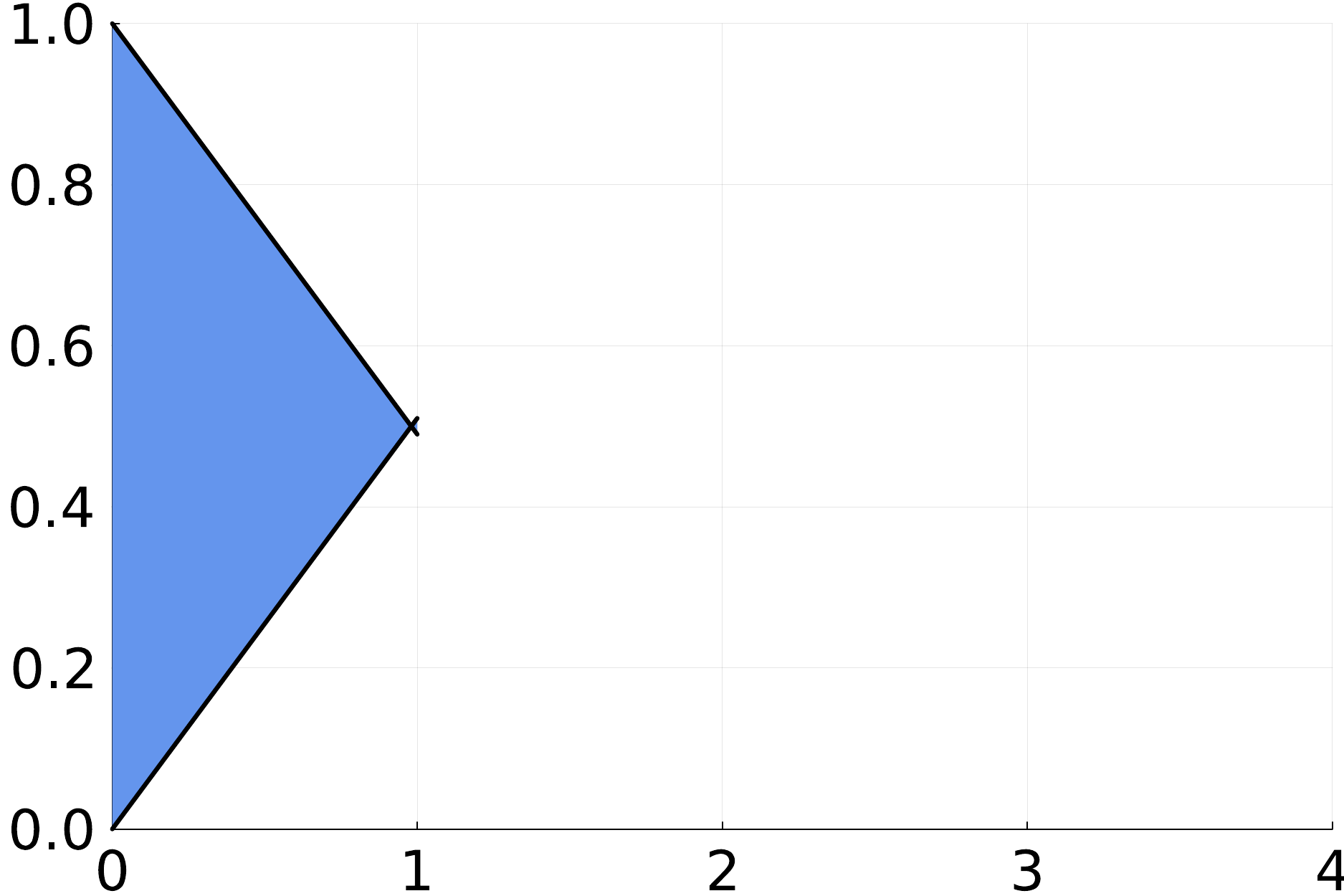}
	\includegraphics[width=0.19\textwidth,keepaspectratio]{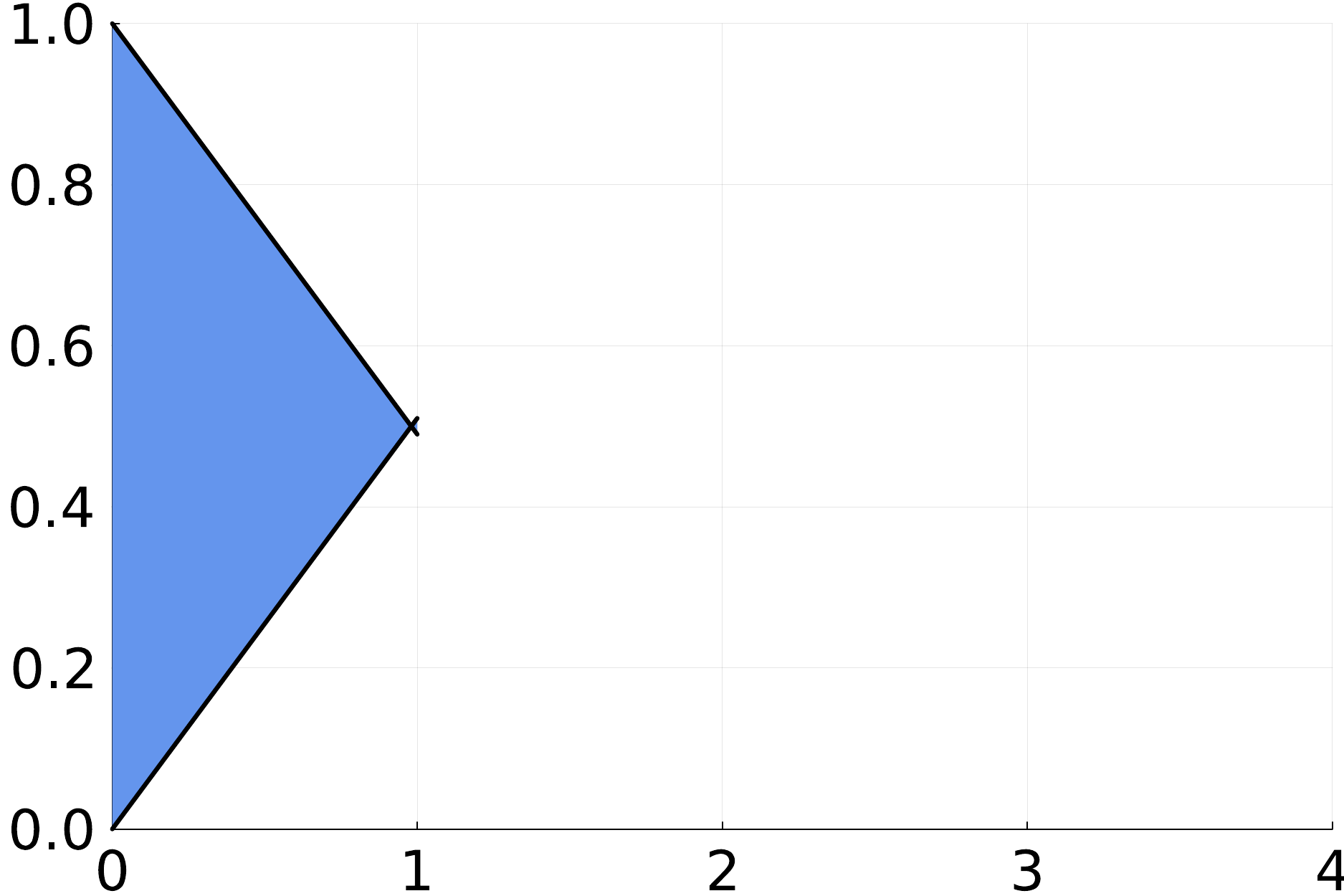}
	\includegraphics[width=0.19\textwidth,keepaspectratio]{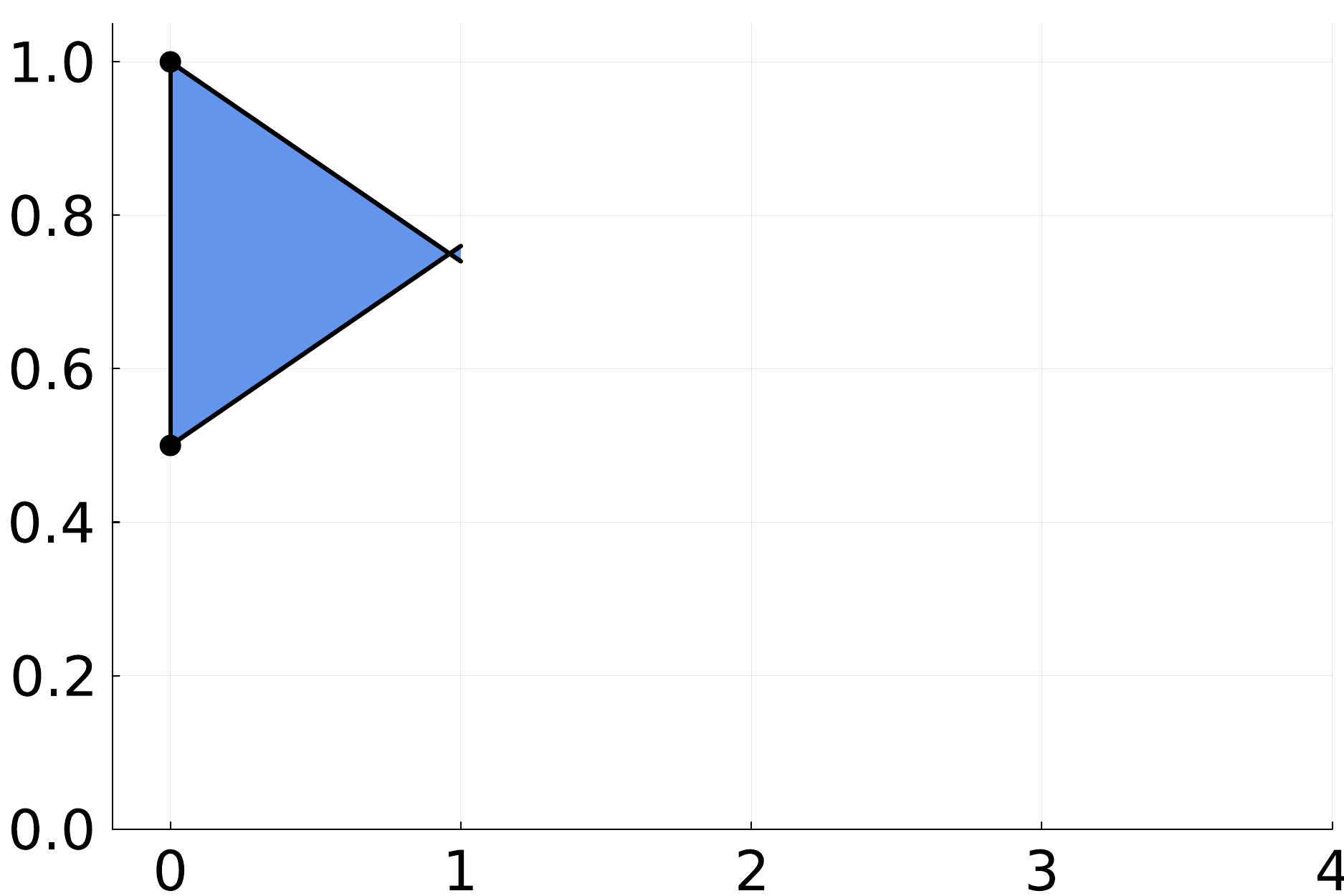}
	
	\caption{Experiment from~\cite{Thrun94} for the DNN in \figref{nn_thrun94}.
	Each row consists of one experiment and each column corresponds to one neuron.
	In the last experiment, the sets become empty after the first iteration, which we represented graphically by converging and crossing bounds.}
	\label{fig:example_thrun94}
\end{figure}

The results are shown in \figref{example_thrun94}.
Each row consists of one experiment and each column corresponds to one neuron.
The horizontal axis shows the number of iterations, with iteration~$0$ being the initial configuration.
The vertical axis shows the range of each neuron valuation.
Input neuron~$1$ is always further constrained to a smaller interval.
In experiments~$1$, $2$, and~$5$, the second input neuron is also further constrained, and in the last three experiments, the output neuron is further constrained.

We can see that both the image and the preimage of the DNN can get refined by forward resp.\ backward propagating bounds.
In the first experiment, we obtain a proof of the following statement:
If $x_1 \in [0, 2]$ and $x_2 \in [0.8, 1]$, then $\NN(x) \in [0.51, 0.79]$.
Note that this can also be achieved with just forward-image computation (the result of which corresponds to the input- and output-neuron bounds after one iteration).
In the third experiment, we obtain a proof of the following statement:
If $x_1 \in [0, 2]$ and $\NN(x) \in [0.5, 1]$, then $x_2 \in [0, 0.41]$ and $\NN(x) \in [0.5, 86]$.
Unlike with just forward propagation, we can derive statements about the inputs, and we also obtained tighter output bounds than what forward-image computation would ($\NN(x) \in [0.5, 0.94]$).
The last experiment finds that the input and output constraints are incompatible, and the sets become empty.
Thus we obtain a proof of the following statement:
Either $x_1, x_2 \notin [0, 2]$ or $\NN(x) \notin [0.5, 1]$.
We refer to~\cite{Thrun94} for further explanation.

\section{Conclusion}

In this paper, we have presented a complete picture of the computation of the preimage of a DNN with piecewise-affine activation functions.
While a similar approach has been presented before~\cite{Maire99}, we believe that it has not been appreciated in the formal-methods community, and we also filled a small technical gap to allow the application to the common class of ReLU networks.
We have discussed applications in interpretability, scalability improvements via over- and underapproximations, and a combined forward and backward computation.

\smallskip

We see many opportunities for future work.
First, the extension to other activation functions has been proposed via piecewise-affine approximation~\cite{Maire99}.
This approximation can also be implemented in a conservative way.
Second, we conjecture that practical abstraction methods can be found, similar to what has been achieved in forward-image computations.
Finally, we envision applications related to robustness and adversarial attacks.
Once we have computed the preimage, we can use it to search for instances of interest in it (e.g., an input that maximizes another objective).
Consider a classifier and the preimage of a small region around a decision boundary.
The preimage contains all instances that will make the decision change with a small perturbation.
We can obtain samples from the preimage set to determine whether the uncertain classification is indeed reasonable.
Also, since we can compute the preimage of all decision boundaries, we can effectively obtain a partition of the input space.

\section*{Acknowledgments}

We thank the anonymous reviewers for helpful comments to improve this paper.
This research was partly supported by DIREC - Digital Research Centre Denmark and the Villum Investigator Grant S4OS.

\bibliographystyle{splncs04}
\bibliography{bibliography}

\end{document}